\documentclass[10pt,journal,compsoc]{IEEEtran}
\usepackage{epsfig}
\usepackage{graphicx}
\usepackage{amsmath}
\usepackage{amssymb}
\usepackage{wasysym}
\usepackage{tabularx}
\usepackage{xcolor}
\usepackage{threeparttable}
\usepackage{algorithm}
\usepackage{algorithmic}
\usepackage{colortbl}
\usepackage{multirow}
\usepackage{graphicx}
\usepackage{subfig}
\usepackage{ragged2e}
\usepackage{bm}
\usepackage[pagebackref=false,breaklinks=true,colorlinks,bookmarks=false]{hyperref}

\usepackage{amsfonts}
\usepackage{array}
\usepackage{textcomp}
\usepackage{stfloats}
\usepackage{url}
\usepackage{verbatim}
\usepackage{booktabs, multirow, hhline}
\usepackage{pifont}
\usepackage{threeparttable}
\usepackage{tcolorbox}
\usepackage{arydshln}
\usepackage{tikz}

\ifCLASSOPTIONcompsoc
  \usepackage[nocompress]{cite}
\else
  \usepackage{cite}
\fi

\ifCLASSINFOpdf
\else
\fi

\usepackage{xspace}
\makeatletter
\newcommand{\thickhline}{%
    \noalign {\ifnum 0=`}\fi \hrule height 0.7pt
    \futurelet \reserved@a \@xhline
}

\newcolumntype{I}{!{\vrule width 0.8pt}}

\DeclareRobustCommand\onedot{\futurelet\@let@token\@onedot}
\def\@onedot{\ifx\@let@token.\else.\null\fi\xspace}

\newcolumntype{P}[1]{>{\centering\arraybackslash}p{#1}}
\definecolor{lightgray}{gray}{.9}
\definecolor{deepgray}{gray}{.8}
\definecolor{mygreen}{RGB}{29, 154, 120}
\definecolor{DarkGreen}{RGB}{42,110,63}

\begin{document}

\title{Multilingual Text-to-Image Person Retrieval via Bidirectional Relation Reasoning and Aligning}

\author{Min Cao, Xinyu Zhou, Ding Jiang, Bo Du,~\IEEEmembership{Senior Member,~IEEE}, Mang Ye,~\IEEEmembership{Senior Member,~IEEE}, \\ 
Min Zhang
\IEEEcompsocitemizethanks{
\IEEEcompsocthanksitem This work is supported by the National Natural Science Foundation of China under Grants 62476188 and 62176188, the Natural Science Foundation of the Jiangsu Higher Education Institutions of China, Key Laboratory of New Generation Artificial Intelligence Technology \& Its Interdisciplinary Applications (Southeast University), Ministry of Education, China (Corresponding author: Mang Ye).

\IEEEcompsocthanksitem Min Cao, Xinyu Zhou, and Min Zhang are with the School of Computer Science and Technology, Soochow University, SuZhou 215000, China (e-mail: mcao@suda.edu.cn, xyzhou2023@stu.suda.edu.cn).

\IEEEcompsocthanksitem Ding Jiang, Bo Du, and Mang Ye are with the School of Computer Science, Wuhan University, Wuhan 430072, China (e-mail: yemang@whu.edu.cn).

 }
}


\markboth{IEEE TRANSACTIONS ON PATTERN ANALYSIS AND MACHINE INTELLIGENCE}%
{Shell \MakeLowercase{\textit{et al.}}: Bare Demo of IEEEtran.cls for Computer Society Journals}

\IEEEtitleabstractindextext{%
\begin{abstract}
\justifying
Text-to-image person retrieval (TIPR) aims to identify the target person using textual descriptions, facing challenge in modality heterogeneity.
Prior works have attempted to address it by developing cross-modal global or local alignment strategies.
However, global methods typically overlook fine-grained cross-modal differences, whereas local methods require prior information to explore explicit part alignments.
Additionally, current methods are English-centric, restricting their application in multilingual contexts.
To alleviate these issues, we pioneer a multilingual TIPR task by developing a multilingual TIPR benchmark, for which we leverage large language models for initial translations and refine them by integrating domain-specific knowledge.
Correspondingly, we propose Bi-IRRA: a Bidirectional Implicit Relation Reasoning and Aligning framework to learn alignment across languages and modalities. 
Within Bi-IRRA, a bidirectional implicit relation reasoning module enables bidirectional prediction of masked image and text, implicitly enhancing the modeling of local relations across languages and modalities, a multi-dimensional global alignment module is integrated to bridge the modality heterogeneity.
The proposed method achieves new state-of-the-art results on all multilingual TIPR datasets. Data and code are presented in https://github.com/Flame-Chasers/Bi-IRRA.
\end{abstract}
}

\maketitle

\IEEEdisplaynontitleabstractindextext

\IEEEpeerreviewmaketitle

\IEEEraisesectionheading{\section{Introduction}\label{sec:intro}}
\IEEEPARstart{G}{iven} a text query, Text-to-Image Person Retrieval (TIPR)~\cite{li2017person} aims to identify the most relevant person images from an extensive gallery of such images. The task is similar to the person re-identification task (Re-ID)~\cite{he2021transreid, luo2019bag, wang2022nformer}, which involves identifying person images across cameras based on the image query. In contrast to the structured image query in Re-ID, the text query in TIPR takes the form of free, flexible characters, making it more accessible and offering substantial application potential in public safety domains. In recent years, TIPR has garnered increasing attention~\cite{jiang2023cross, bai2023rasa, cao2024empirical, zuo2024ufinebench, tan2024harnessing}.

\begin{figure}[tbp]
\includegraphics[width=0.48\textwidth]{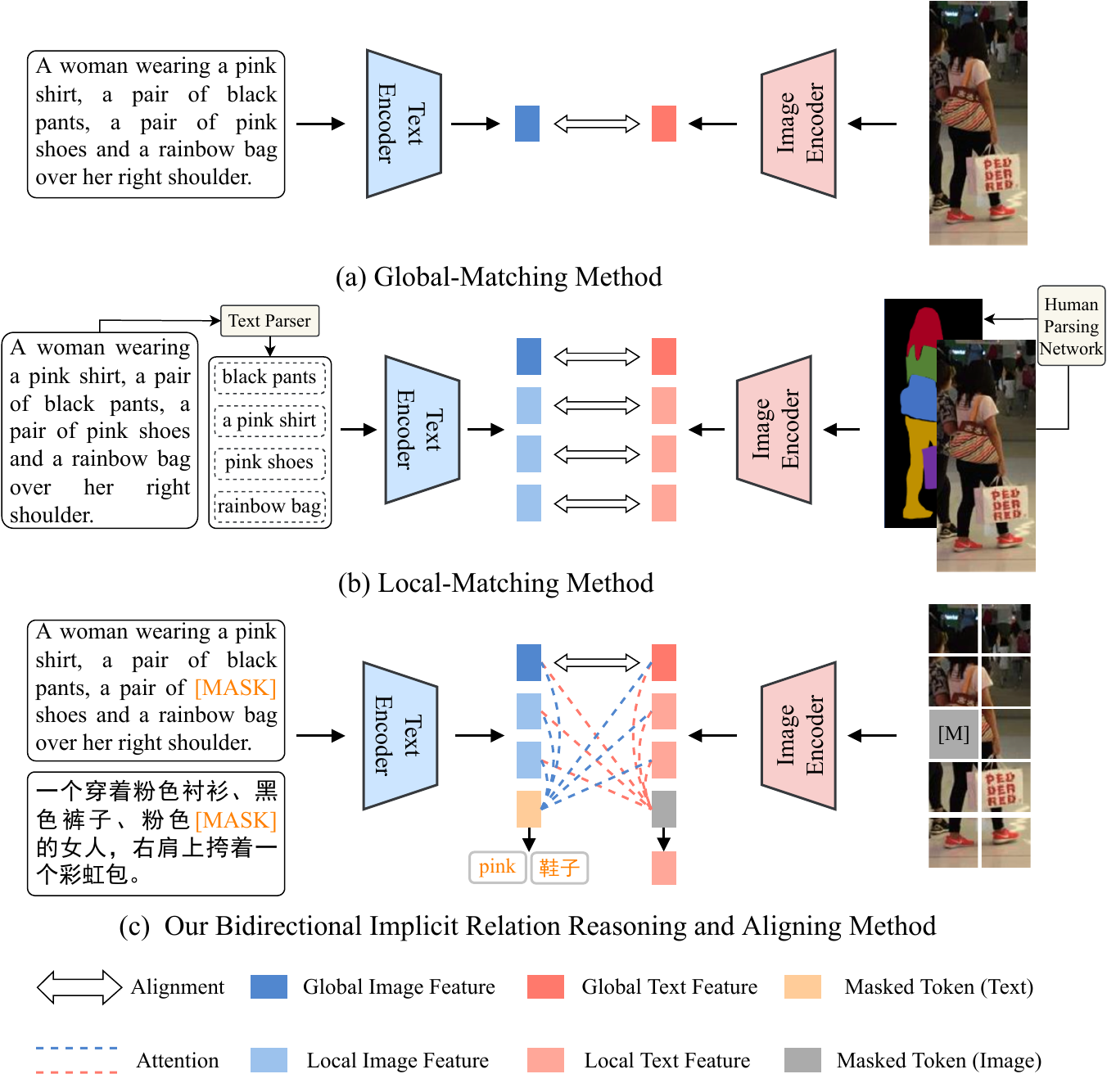}
\caption{Illustration of different TIPR methods. (a) Global-matching methods directly align global image and text feature representations. (b) Local-matching methods explicitly extract and align local image and text representations. (c) Our bidirectional implicit relation reasoning and aligning method not only implicitly reasons about the relations among all local tokens but also aligns global image and text representations in a multilingual environment.}
\label{fig: intro}
\end{figure}

A key challenge in TIPR is the inherent modality gap between vision and language, driving research toward robust cross-modal alignment.
These efforts can be broadly categorized into global-matching~\cite{zhang2018deep, zheng2020dual, chen2018improving, wang2019language}\cite{shu2022see,wu2023refined,liu2025dm} and local-matching methods~\cite{chen2022tipcb, ding2021semantically, jing2020pose, wang2020vitaa}\cite{you2025diverse}. 
The former aligns global text-image representations at the coarse-grained level via cross-modal matching loss functions (Fig.~\ref{fig: intro}(a)), while the latter establishes fine-grained associations between textual entities and image body parts (Fig.~\ref{fig: intro}(b)).

Despite notable progress in this task, two critical issues remain to be addressed. The first issue is the limited focus on language environments. Current methods~\cite{zuo2024ufinebench, jiang2023cross, bai2023rasa, wu2024laip, cao2024empirical} center around addressing TIPR with English as the default text query. However, in practical application, there is a growing need for supporting multiple languages, (\emph{e.g.}, Chinese and French) as queries for TIPR. Biased towards a single language (\emph{i.e.}, English), current methods struggle to achieve accurate retrieval when confronted with diverse language requirements in real-world scenarios.
Another issue involves the constraint of existing cross-modal alignment strategies. 
Global-matching methods align the global representations of texts and images, often overlooking the need to bridge modality heterogeneity at a more detailed, fine-grained level, potentially impacting performance.
On the other hand, local-matching methods are tailored to build the explicit correspondence between body parts and textual descriptions with the aid of external technologies and predefined rules.
As a result, they confine the exploration of cross-modal alignment within the boundaries of these set rules, and also pose resource-intensive demands as it necessitates the extraction and storage of multiple local part representations of images and texts during inference. 

In this paper, we pioneer a multilingual TIPR task. Specifically, we build the multilingual TIPR benchmark and propose Bi-IRRA: a cross-modal Bidirectional Implicit Relation Reasoning and Aligning framework to learn alignment across languages and modalities at both coarse-grained and fine-grained levels. This work is centered on addressing both data and framework aspects.

\textbf{Data}. For a novel multilingual TIPR task, a critical obstacle is the scarcity of multilingual TIPR data to support its research. While manual annotation of multilingual TIPR data presents a direct solution, it is labor-intensive and impractical for covering a wide range of languages over extensive image data. 
An alternative solution involves utilizing Large Models (LMs)~\cite{dubey2024llama, ouyang2022training, bai2023qwenvl, abdin2024phi, zhu2023minigpt, liu2024visual} for automatic translation on existing TIPR datasets, thereby extending them beyond English. 
However, directly using LMs for translation often introduces noise due to their lack of domain-specific knowledge. For this, we develop a LMs-driven Domain Adaptive Translation (LDAT) pipeline, consisting of translation, filtering, and rewriting phases. After an initial translation by Large Language Models (LLMs)~\cite{bai2023qwen, dubey2024llama, ouyang2022training} in the translation phase, we identify clean and noisy translation texts in the filtering phase. Subsequently, the clean texts with corresponding person images serve as the supervision signal to finetune Multimodal Large Language Models (MLLMs)~\cite{bai2023qwenvl, abdin2024phi, zhu2023minigpt}. The finetuned MLLMs, enriched with comprehensive domain-specific knowledge, are employed to rewrite noisy texts in the rewriting phase, thus effectively mitigating the noise issue. The proposed LDAT, as a concise translation pipeline, enables the cost-effective acquisition of the high-quality multilingual TIPR benchmark.

\textbf{Framework}. 
In contrast to the traditional TIPR task that deals solely with heterogeneity between text and image modalities, the multilingual TIPR encapsulates modality heterogeneity and linguistic diversity challenges.
In response, we propose the Bi-IRRA framework, designed to establish robust global alignment and explore implicit fine-grained relations across diverse languages and modalities (as depicted in Fig.~\ref{fig: intro} (c)).
Specifically, Bi-IRRA comprises a Bidirectional Implicit Relation Reasoning (Bi-IRR) module and a Multi-dimensional Global Alignment (Md-GA) module.
The Bi-IRR module performs bidirectional prediction of masked text and image, enabling the implicit modeling of local relations between vision and language.
It includes a \emph{bi-lingual Masked Language Modeling (MLM)} pretext task and a \emph{cross-lingual Distillation Masked Image Modeling (D-MIM)} pretext task, tailored for adaptive modeling and interaction across languages.
Meanwhile, Md-GA aligns global text and image representations from multiple dimensions, facilitating the extraction of discriminative global text and image representations.
The Md-GA module comprises a \emph{bi-lingual Image-Text Contrastive (ITC)} pretext task and a \emph{bi-lingual Asymmetric Image-Text Matching (A-ITM)} pretext task, with the former applied to the unimodal encoders and the latter to the subsequent multimodal interaction encoder.
Notably, an asymmetric masking operation on input data is integrated into the \emph{bi-lingual A-ITM} to facilitate noise-robust learning for noisy target text.
Finally, by integrating these two modules, Bi-IRRA achieves comprehensive alignment across languages and modalities at both fine-grained and coarse-grained levels.

Our main contributions are as follows.
1) We pioneer a multilingual TIPR task that enables querying in multiple languages to retrieve the target person, offering substantial real-world application potential compared to traditional TIPR.
2) We develop the LDAT pipeline to acquire multilingual TIPR data. By introducing domain-specific knowledge into LMs to mitigate the noise issue in LMs-driven translation, LDAT enables the construction of high-quality multilingual TIPR benchmarks.
3) To address both modality heterogeneity and linguistic diversity in multilingual TIPR, we propose the Bi-IRRA framework, which learns bidirectional implicit relations at a fine-grained level while attaining global alignment at a coarse-grained level across languages and modalities.
4) Extensive experiments on the multilingual TIPR benchmarks demonstrate that the proposed framework surpasses existing SOTA methods.


A preliminary version of this work has been published in CVPR 2023~\cite{jiang2023cross}. This paper presents the following improvements. 
(1) We introduce a more practical multilingual TIPR task that extends beyond the former traditional TIPR task. To support this, we develop the LDAT pipeline to construct high-quality multilingual TIPR benchmarks cost-effectively.
(2) To adapt the framework for multilingual TIPR, we introduce key improvements to the framework. 
Firstly, we integrate a Bi-IRR module into the framework. Going beyond the original Implicit Relation Reasoning (IRR) module, which only featured a MLM pretext task, we introduce a novel \emph{cross-lingual D-MIM} pretext task. It aids in information reconstruction in the image domain by facilitating cross-lingual relations. 
We also refine the original MLM into a \emph{bi-lingual MLM}, specifically designed for multilingual modeling.
Significantly, Bi-IRR, combined with these two pretext tasks, enables bidirectional implicit relation reasoning for both masked textual and visual content, substantially enhancing cross-modal fine-grained alignment capabilities.
Secondly, we restructure the global alignment across modalities. 
In previous work, the Similarity Distribution Matching (SDM) pretext task was employed for global alignment. It only constrains unimodal encoders to generate separate global representations for image and text, lacking deep cross-modal interaction and fusion.
In contrast, we develop a Md-GA module to align global representations across languages and modalities from multiple dimensions.
Notably, we also consider the noisy interference from the generated target texts during global alignment by designing a \emph{bi-lingual A-ITM} pretext task into Md-GA.
(3) We expand experiments to encompass a wider range of language environments for TIPR, accompanied by more thorough analyses.
To the best of our knowledge, our work is the first effort to address TIPR in a multilingual setting, significantly promoting its practical application value.

\section{Related Work}
\subsection{Text-to-Image Person Retrieval}
Since Li \textit{et al.}~\cite{li2017person} pioneered the TIPR task, there have been notable advancements in its development~\cite{jiang2023cross, cao2024empirical, lin2024cross, hu2024personmae, zuo2024ufinebench}. The key challenge in TIPR lies in the significant modality heterogeneity between vision and language. Existing methods address this challenge by focusing on representation learning and cross-modal alignment.

\textbf{Representation Learning}.
This category of methods is centered on the development of robust representation learning network designed to extract discriminative feature representations relevant to individuals.
Early works~\cite{chen2018improving, li2017identity, li2017person, chen2021cross, sarafianos2019adversarial} employed convolutional neural networks~\cite{simonyan2014very, he2016deep} as image encoder and LSTM~\cite{hochreiter1997long} as text encoder. Later works~\cite{li2022learning, zhang2018deep} improved these networks with ViT~\cite{dosovitskiy2020image} for images and BERT~\cite{kenton2019bert} for texts. More recent advancements have embraced powerful Vision-Language Pre-training (VLP) models~\cite{radford2021learning, li2021align} (\emph{e.g.}, CLIP~\cite{radford2021learning}) as the representation learning networks. These VLP models are typically pre-trained on large-scale vision-language datasets, enabling them to extract more discriminative person feature representations, thereby garnering considerable attention~\cite{jiang2023cross, bai2023rasa, tan2024harnessing, wu2024text, cao2024empirical}.
For instance, Han \textit{et al.}~\cite{han2021text} introduced the CLIP model into TIPR, devising a momentum contrastive learning framework to transfer knowledge from large-scale image-text pairs to the representation learning in TIPR; Cao \textit{et al.}~\cite{cao2024empirical} conducted a comprehensive empirical study of CLIP for TIPR, establishing a robust TBPS-CLIP baseline for effective representation learning in this task.

\textbf{Cross-modal Alignment}.
The second category of methods is dedicated to designing an effective alignment strategy to achieve favorable cross-modal alignment.
Global-matching methods~\cite{zhang2018deep, zheng2020dual, chen2018improving, wang2019language, cao2024empirical} aligned global textual and visual representations directly through designing the rational cross-modal matching loss functions. While these methods are straightforward and intuitive, they often overlook fine-grained information when performing cross-modal alignment.
Later, local-matching methods~\cite{niu2020improving, wang2022caibc, wang2022look, wu2024text, wang2020vitaa, wu2021lapscore, jing2020pose} have been introduced to align fine-grained visual and textual information, enhancing cross-modal alignment.
Typically, most of these methods~\cite{wang2020vitaa, wu2021lapscore, jing2020pose, chen2022tipcb, fujii2023bilma} relied on external technologies and predefined rules to explicitly extract local textual and visual information, such as text phrases, human segmentation~\cite{wang2020vitaa, fujii2023bilma}, and color information~\cite{wu2021lapscore}, to model fine-grained relations between two modalities. For example, Fujii \textit{et al.}~\cite{fujii2023bilma} leveraged human parsing models to obtain semantic labels of images, which serve as supervision signals to align cross-modal information.
Although incorporating such fine-grained information enhances retrieval performance, these explicit local-matching methods introduce additional computational complexity during inference when computing the similarity of all these local part representations of images and texts.
In comparison, several implicit local-matching methods~\cite{jiang2023cross, bai2023rasa, yan2023image, yang2023towards, li2024adaptive, zuo2024ufinebench} have been proposed. These methods explore fine-grained cross-modal alignment relations without relying on external explicit dependencies, significantly reducing additional computational overhead. For example, the conference version of this work~\cite{jiang2023cross} leveraged the MLM pretext task where visual and unmasked textual information are integrated to predict the masked textual tokens. These masked tokens are treated as anchors to align fine-grained cross-modal information implicitly. 

Despite progress made in TIPR, existing methods are predominantly limited to the monolingual English TIPR.
Unlike these methods, this work pioneers the exploration of the multilingual TIPR task and proposes the Bi-IRRA framework, which performs implicit modeling of fine-grained information across different languages and modalities.

\begin{figure*}[t]
\centering
\includegraphics[width=\textwidth, height=0.3\textwidth]{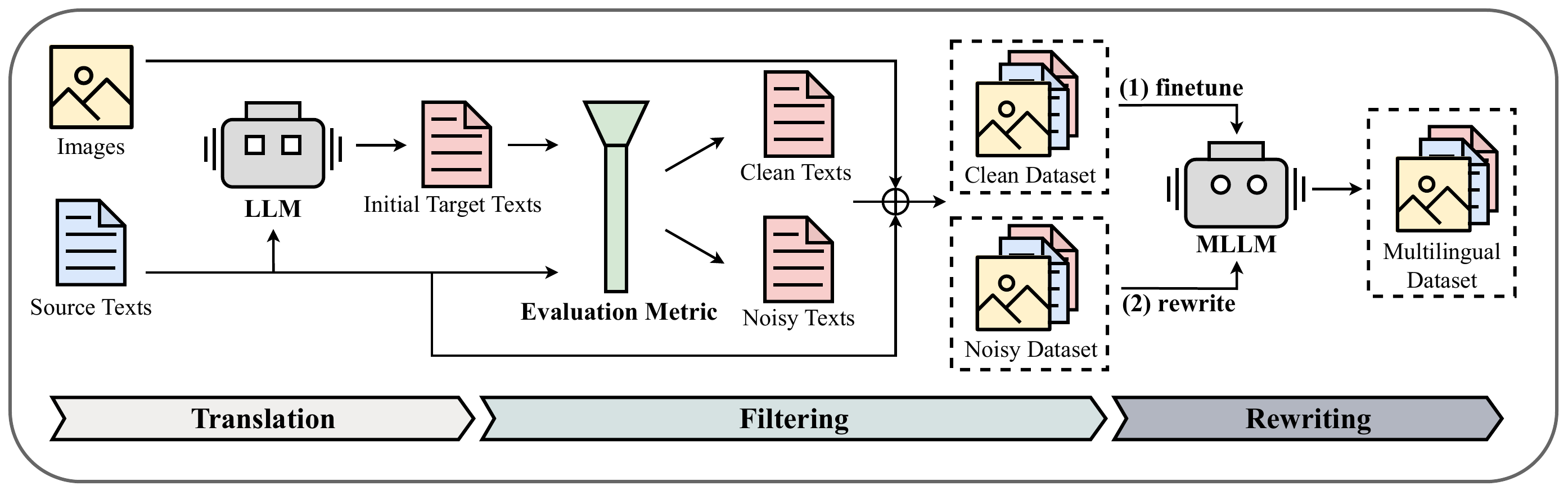}
\caption{An Overview of LMs-driven Domain Adaptive Translation (LDAT) pipeline. It consists of three main phases: translation, filtering, and rewriting. In the translation phase, the source texts are translated into the initial target texts by an LLM. During the filtering phase, an evaluation metric is adopted to assess the noise level of each initial target text, based on which we divide the initial target texts into two groups: clean texts and noisy texts. These texts are then paired with their corresponding images and source texts, resulting in a clean dataset and a noisy dataset, respectively. The rewriting phase includes two steps: (1) finetuning an MLLM with the clean dataset, and (2) using the finetuned MLLM to rewrite the initial target texts from the noisy dataset, thereby producing a high-quality multilingual TIPR dataset.}
\label{fig: pipeline}
\end{figure*}

\subsection{Multilingual Image-Text Retrieval}
Multilingual Image-Text Retrieval (MITR)~\cite{elliott2016multi30k, portaz2019image, aggarwal2020towards, nie2024improving, wang2024multimodal} involves achieving image-text retrieval across multiple languages. It emphasizes retrieval on common instances rather than concentrating solely on individual instances as in multilingual TIPR.
To tackle this task, learning feature representations from large-scale multilingual vision-language datasets is straightforward and efficient. Consequently, constructing such large-scale multilingual vision-language datasets has become a key focus in recent MITR research~\cite{ni2021m3p, jain2021mural, li2023unifying, zhou2021uc2}.

Two kinds of methods are typically used for the construction of datasets. 
The first method~\cite{ni2021m3p, jain2021mural, zeng2022cross, li2023unifying} integrated existing English-centric vision-language datasets with additional multilingual parallel text corpora. Although these datasets can be scaled up with minimal human effort, they lack direct alignments of non-English text and image pairs.
The second method~\cite{zhou2021uc2} involved leveraging machine translation to automatically extend existing vision-language datasets beyond English. This results in new datasets with direct alignments of images and texts across all languages, and yet introduces noise from machine translation.

Building upon these constructed datasets, some research efforts focus on designing alignment strategies to bridge different languages and modalities. For works trained on datasets built by the first approach~\cite{fei2021cross, ni2021m3p, carlsson2022cross, jain2021mural, li2023unifying}, they typically model cross-modal and cross-lingual alignments separately, using English texts as a pivot to align images with non-English texts indirectly. For works trained on datasets built by the second approach~\cite{zhang2022multi, wang2022cross, wang2024dual, wang2024cl2cm, cai2024cross}, they tend to fully exploit the inherent correspondences between languages and modalities to minimize noise during model training.
For example, Cai \textit{et al.}~\cite{cai2024cross} employed a knowledge distillation mechanism to extract effective information from non-English texts with the aid of English texts, thereby achieving the robust and solid alignment between different languages and modalities.

Different from these MITR methods, our work contributes to a multilingual TIPR task, which needs to model more fine-grained information for individuals across languages and modalities.
Given the domain specificity of this task, we propose the LDAT pipeline, which incorporates domain-specific knowledge to construct high-quality multilingual data. 
In terms of framework design, we place a greater emphasis on fine-grained cross-modal alignment by introducing the Bi-IRRA framework.

\section{LMs-Driven Domain Adaptive Translation} 
\label{sec:data}

Beginning with existing TIPR dataset $\mathcal{D}=\left\{I_n, T^s_n \right\}^{N}_{n=1}$, containing the \(n\)-th person image $I_n$ and its corresponding English text $T^s_n$ (referred to as the source text), we build its multilingual counterpart $\mathcal{D}_\mathcal{M}=\left\{I_n, T^s_n, T^t_n \right\}^{N}_{n=1}$, where \( T^t_n\) represents the non-English text (referred to as the target text) and is paired with \(I_n\) and \(T^s_n\).

We propose the LDAT pipeline\footnote{For clarity and simplicity in presentation, we omit the subscript \( n \) from \( I_n, T^s_n, T^t_n \) when it is not essential in the following description.} to automatically acquire $T^t$.
First, an LLM is employed to translate each source text \(T^s\) into its corresponding initial target text \( \widetilde{T}^t \). 
These initial target texts are then filtered into clean and noisy texts. 
Finally, noisy texts are rewritten by the MLLM finetuned on domain-specific data, which includes clean texts along with the person images, to produce the final high-quality target texts.
To sum up, this process, depicted in Fig.~\ref{fig: pipeline}, comprises three key phases: translation, filtering, and rewriting.

\textbf{Translation}. 
We first translate the source text $T^s$ in English, to the target text $\widetilde{T}^t$ in a non-English language, by leveraging the language understanding capability of LLMs. Specifically, we launch an LLM~\cite{bai2023qwen, dubey2024llama} with a fixed instruction template:

\begin{tcolorbox}
\textit{Please translate the English sentence `\{source-text\}' into \{target-language\}}. 
\end{tcolorbox}

Here, the placeholder \textit{\{source-text\}} is filled with the source text \(T^s\), and \textit{\{target-language\}} is replaced with the target language name, (\emph{e.g.}, Chinese or French). With this simple yet efficient process, we obtain the initial target text \(\widetilde{T}^t\) corresponding to the source text \(T^s\), as shown in Fig~\ref{fig: example display}. 
Since the source texts describing individuals typically follow a consistent sentence structure, the powerful LLM can attain relatively accurate translations, as shown in Fig.~\ref{fig: example display} (a) \(\sim\) (c). 
However, due to the lack of domain-specific knowledge, LLM still inevitably introduces a small amount of noise during the translation process, which could potentially impact the subsequent training process in multilingual TIPR. 

\textbf{Filtering.}
To mitigate the noise issue, we implement a filtering phase to classify the initial target texts into two groups: clean texts and noisy texts. The classification facilitates the rewriting of noisy texts during the subsequent rewriting phase.

Fig.~\ref{fig: example display} (d) \(\sim\) (i) showcases several specific examples of noisy texts.
For example, in Fig.~\ref{fig: example display} (d), the LLM incorrectly translates the phrase \textit{`A man wearing white shoes on his feet'} from the source text to \textit{`A man in white shoes is on his feet'} in Chinese. 
The inherent hallucination~\cite{huang2023survey} in the LLM could lead to such noisy translations, and its lack of domain-specific knowledge further exacerbates the hallucination.

Therefore, we assess the translation quality of each initial target text as an indicator of its noise level.
Conventional machine translation metrics like BLEU~\cite{papineni2002bleu} and METEOR~\cite{banerjee2005meteor} typically necessitate reference translations for evaluation, such references are unavailable for our task, making these metrics unsuitable. Hence, we develop a reference-free machine translation evaluation metric to serve as the noise level indicator. Specifically, we employ a multilingual pre-trained language model, COMETWiki~\cite{rei2022cometkiwi}, to extract linguistic representations of \(n\)-th source text \(T^s_n\) and initial target text \(\widetilde{T}^t_n\). Then the translation quality is evaluated by measuring the similarity $\Phi(T^s_n, \widetilde{T}^t_n)$ between their extracted representations.
The noise level $\phi_n$ of \(n\)-th initial target text \(\widetilde{T}^t_n\) is calculated as follows:
\begin{equation} \label{eq:1}
\phi_n = 1-\Phi(T^s_n, \widetilde{T}^t_n).
\end{equation}

We introduce a threshold \(\theta\) to divide these initial target texts into clean and noisy texts. Specifically, if \(\phi_n \leq \theta\), the corresponding \(\widetilde{T}^t_n\) is classified as the clean text and also serves as the final target text \(T^t_n\); otherwise, it is identified as the noisy text. Subsequently, by pairing these texts with their respective images and source texts, we construct two datasets: a clean dataset \(\mathcal{D}^C_{\mathcal{M}}\) and a noisy dataset \(\mathcal{D}^N_{\mathcal{M}}\):
\begin{equation}
\label{eq: 2}
\mathcal{D}^C_{\mathcal{M}} = \{(I_n, T^s_n, T^t_n) \mid \phi_n \leq \theta \}^{N_{clean}}_{n=1},
\end{equation}
\begin{equation}
\label{eq: 3}
\mathcal{D}^N_{\mathcal{M}} = \{(I_n, T^s_n, \widetilde{T}^t_n) \mid \phi_n > \theta \}^{N_{noisy}}_{n=1}.
\end{equation}
Here, \(N_{clean}\) and \(N_{noisy}\) are the numbers of clean texts and noisy texts, respectively, and \(N_{clean} + N_{noisy} = N\).

\textbf{Rewriting.}
The goal of this phase is to rewrite the noisy text \(\widetilde{T}^t\) from \(\mathcal{D}^N_{\mathcal{M}}\). 
We first utilize \(\mathcal{D}^C_{\mathcal{M}}\) as the supervision signal to finetune an MLLM~\cite{bai2023qwenvl, abdin2024phi} and then employ the finetuned MLLM to execute the rewriting process.

Specifically, for each triplet data \((I, T^s, T^t) \in \mathcal{D}^C_{\mathcal{M}}\), we construct an instruction based on a predefined template:

\begin{tcolorbox}
\textit{\textless img\textgreater\{image-path\}\textless/img\textgreater \ Please combine the image information, translate the English sentence `\{source-text\}' into \{target-language\}}. 
\end{tcolorbox}

In this template, \textit{\textless img\textgreater} and \textit{\textless/img\textgreater}\ serve as the special tokens to indicate the region of image input. The placeholder \textit{\{image-path\}} is replaced with the path to the image \(I\), which enables the model to access and process visual information. \textit{\{source-text\}} and \textit{\{target-language\}} are substituted with \(T^s\) and the target language name, respectively. 

\begin{figure*}[ht]
\centering
\includegraphics[width=\textwidth]{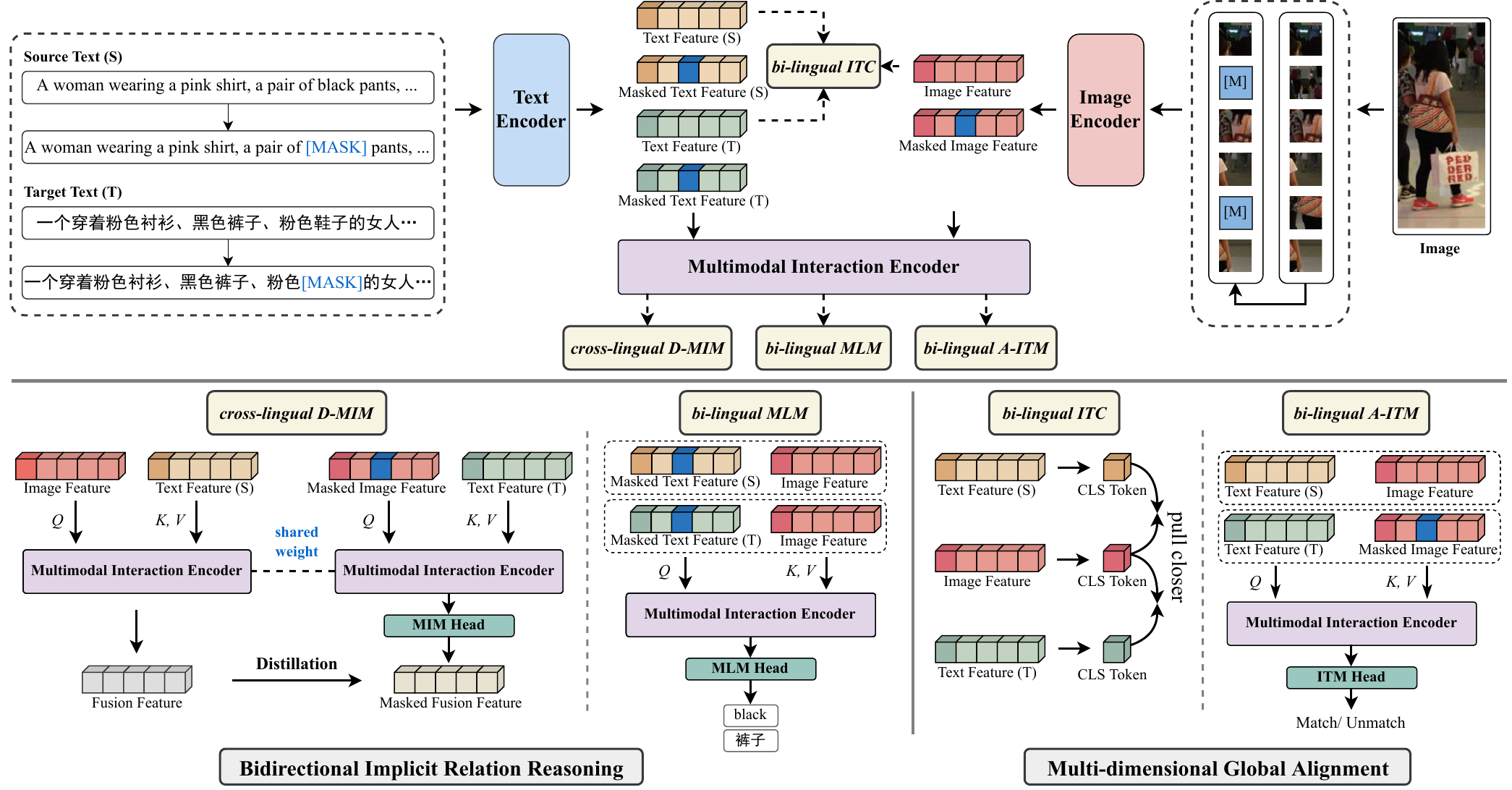}
\caption{Overview of the proposed Bidirectional Implicit Relation Reasoning and Aligning (Bi-IRRA) framework. It consists of an image encoder, a text encoder, and a multimodal interaction encoder. With the triplet data as input, Bi-IRRA achieves cross-lingual cross-modal alignment through two key modules. The first is the Bidirectional Implicit Relation Reasoning (Bi-IRR) module, which has two pretext tasks\textemdash \emph{cross-lingual D-MIM} and \emph{bi-lingual MLM}. Bi-IRR facilitates the bidirectional modeling of fine-grained relations across languages and modalities. The second is the Multi-dimensional Global Alignment (Md-GA) module, which contains \emph{bi-lingual ITC} and \emph{bi-lingual A-ITM} pretext tasks to align global feature representations of texts and images.}
\label{fig:method}
\end{figure*}

We use the constructed instruction as input, with the target text \(T^t\) serving as the ground truth, to finetune an MLLM with LoRA~\cite{hu2021lora} in a fully supervised manner. Subsequently, the same template employed during the finetuning process is used by incorporating the image \(I\) and the source text \(T^s\) from \(\mathcal{D}^N_{\mathcal{M}}\), constructing the instruction to guide the finetuned MLLM to re-translate source text into the target text. These re-translated target texts with their corresponding images and source texts are added to \(\mathcal{D}^C_{\mathcal{M}}\), ultimately forming a complete multilingual dataset \(\mathcal{D}_\mathcal{M} = \{I_n, T^s_n, T^t_n\}_{n=1}^N\).

Typically, finetuning the MLLM with high-quality person data from the clean dataset \(\mathcal{D}^C_{\mathcal{M}}\) enriches it with the domain-specific knowledge, which helps alleviate noise from the noisy dataset \(\mathcal{D}^N_{\mathcal{M}}\). 
As illustrated in Fig.~\ref{fig: example display} (d) \(\sim\) (i), the errors in the initial target texts are accurately corrected after rewriting.

\section{Cross-modal Bidirectional Implicit Relation Reasoning and Aligning}

In this section, we elaborate on the proposed Bi-IRRA framework. The overview of Bi-IRRA is illustrated in Fig.~\ref{fig:method} and the details are presented in the following subsections.

\subsection{Architecture} \label{sec:arch}
The Bi-IRRA framework consists of an image encoder, a text encoder, and a multimodal interaction encoder. The image and text encoders each consist of \(12\)-layer transformer blocks, tailored to handle unimodal information. The multimodal interaction encoder, made up of \(6\)-layer transformer blocks, enables cross-modal interaction between image and text through the cross-attention mechanism. 

Bi-IRRA takes the triplet data \((I, T^{s}, T^{t})\) as input for these encoders. Given the image \(I\), it is first divided into \(M\) non-overlapping patches and fed into the image encoder to obtain a sequence of image representations \(F(I) = \{v_{cls}, v_1, \cdots, v_{M} \}\), where \(v_{cls}\) is the global representation and \(v_i\) \((i=1, \cdots, M)\) represents the \(i\)-th patch representation. For the source text \(T^{s}\), the text encoder extracts a sequence of token representations \(F(T^{s}) = \{s_{cls}, s_1, \cdots, s_{L}\}\), where \(s_{cls}\) is the global representation and \(s_i\) \((i=1, \cdots, L)\) donates \(i\)-th token representation, with \(L\) being the number of textual tokens. Similarly, the target text \(T^{t}\) is fed into the text encoder to obtain \(F(T^{t}) = \{t_{cls}, t_1, \cdots, t_{L}\}\). Finally, these unimodal feature representations \(F(I)\), \(F(T^s)\), and \(F(T^t)\) are fed into the multimodal interaction encoder to generate a sequence of fusion feature representations.

To effectively encode these representations to achieve robust alignment across languages and modalities, Bi-IRRA is equipped with two core modules: the Bidirectional Implicit Relation Reasoning (Bi-IRR) and Multi-dimensional Global Alignment (Md-GA) modules. 
The Bi-IRR module leverages the \emph{bi-lingual Masked Language Modeling (MLM)} and \emph{cross-lingual Distillation Masked Image Modeling (D-MIM)} on fusion representations to implicitly capture local relations across different languages and modalities.
The Md-GA module employs the \emph{bi-lingual Image-Text Contrastive (ITC)} on unimodal representations and \emph{bi-lingual Asymmetric Image-Text Matching (A-ITM)} on fusion representations to achieve the alignment of global representations between image and text.

\subsection{Bidirectional Implicit Relation Reasoning}
It is crucial to mitigate the modality heterogeneity between vision and language. 
For this, we propose the Bi-IRR module, composed of \emph{cross-lingual D-MIM} and \emph{bi-lingual MLM}, to implicitly align local representations across various languages and modalities through the reconstruction of masked data contents.
The \emph{cross-lingual D-MIM} establishes fine-grained relations by reconstructing masked image information, while \emph{bi-lingual MLM} achieves this by reconstructing masked text information. This bidirectional relation reasoning strengthens the interactions between vision and language in a multilingual scenario.

\textbf{Cross-Lingual Distillation Masked Image Modeling.} 
We propose the \emph{cross-lingual D-MIM} pretext task to reconstruct masked image data at the feature level with the aid of available visual and textual information through a cross-lingual distillation mechanism.
More precisely, considering that the source text generally exhibits higher quality than the target text, and cross-modal learning between the source text and image is more robust than that between the target text and image, we distill the fusion feature representations of the source text and image (as teacher) into the reconstruction of masked image data based on unmasked image data and target textual information (as student). 


Given the image \(I\), we randomly mask a part of image patches with the probability \(p_{\text{img}}\) by applying a blockwise masking strategy~\cite{bao2021beit}, where the contiguous patches are masked and replaced by a learnable masked token, resulting in a masked image denoted as \(\hat{I}\). Correspondingly, the masked image feature \(F(\hat{I})\) is obtained by feeding \(\hat{I}\) into the image encoder. The multimodal interaction encoder with the input of \(F(I)\) and \(F(T^s)\) is used as the teacher model, while that with the input of \(F(\hat{I})\) and \(F(T^t)\) is used as the student model. 
Specifically, taking the teacher model as an example, the multimodal interaction encoder performs the cross-modal fusion via image representation \(F(I)\) as query ($\mathcal{Q}$) and text representation \(F(T^s)\) as key ($\mathcal{K}$) and value ($\mathcal{V}$) and we obtain the fusion representations:
\begin{equation}
G(I, T^{s}) = Transformer(softmax(\frac{\mathcal{Q}\mathcal{K}^{\mathsf{T}}}{\sqrt{d}})\mathcal{V}),
\end{equation}
\begin{equation}
\mathcal{Q} = F(I)W^{\mathcal{Q}}, \quad \mathcal{K} = F(T^s)W^{\mathcal{K}}, \quad \mathcal{V} = F(T^s)W^{\mathcal{V}},
\end{equation}
where $W^{\mathcal{Q}}$, $W^{\mathcal{K}}$ and $W^{\mathcal{V}}$ are the trainable parameter matrices, and $d$ is the representation dimension.
Similarly, we can obtain the fusion representations $G(\hat{I}, T^{s})$ from the student model.
Notably, the multimodal interaction encoders in the teacher model and student model share weights.
Although the encoder processes two distinct inputs—image paired with source text and image paired with target text—both inputs describe the same information and thus share equivalent semantic meaning. By using a shared encoder, we encourage the model to learn a unified multimodal representation space, where semantically similar image-text pairs, regardless of language, are mapped close together. This design inherently promotes semantic consistency and facilitates alignment across languages.

Then, a cross-lingual D-MIM head \(\Psi_{d\text{-}mim}\), implemented as a multi-layer perception, reconstructs the masked image at the feature level based on the fusion representations \(G(\hat{I}, T^t)\) from the student model, resulting in \(\Psi_{d\text{-}mim}(\hat{I}, T^t)\). Then, the \emph{cross-lingual D-MIM} is computed:
\begin{equation}
\mathcal{L}_{d\text{-}mim} = \mathbb{E}_{(I, T^{s}, T^{t}) \sim \mathcal{D}_\mathcal{M}} \mathcal{C}(G(I, T^{s}), \Psi_{d\text{-}mim}(\hat{I}, T^{t})),
\end{equation}
where the fusion representations \(G(I, T^{s})\) from the teacher model serve as supervision, $\mathcal{C}$ represents the cosine similarity. 
It is worth noting that during the optimization process, the fusion representations \(G(I, T^{s})\) do not participate in the backward gradient propagation to avoid model collapse.

\begin{figure}[tbp]
\includegraphics[width=0.45\textwidth, height=0.27\textwidth]{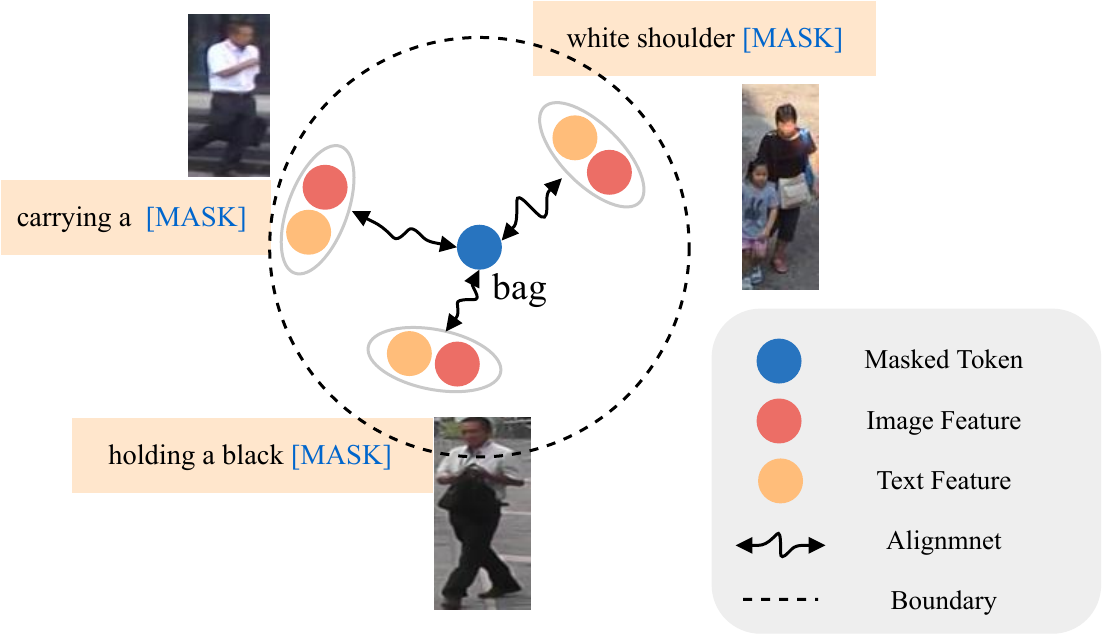}
\caption{Illustration of \emph{bi-lingual MLM}, as exemplified with English text-image pairs. It uses masked textual tokens as local fine-grained keys to align visual and textual information.}
\label{fig: mlm}
\end{figure}

\textbf{Bi-Lingual Masked Language Modeling.} 
We develop the \emph{bi-lingual MLM} to predict masked tokens in both the source and target texts, using unmasked textual and visual information.
The masked textual tokens act as anchors to align the image and text representations, as shown in Fig.~\ref{fig: mlm}. 

Specifically, for the source text \( T^{s} \), we randomly mask each textual token with a probability \(p_{txt}\), replacing the masked tokens with the special token [MASK]. The masked text, denoted as \( \hat{T}^{s} \), is fed into the text encoder to obtain the masked source text representations \(F(\hat{T}^{s})\). 
The multimodal interaction encoder fuses \( F(I) \) and \( F(\hat{T}^{s}) \), producing the fusion representations.
In contrast to the cross-attention fusion
in \emph{cross-lingual D-MIM}, the cross-modal fusion performed here employs text representations as queries and text representations as keys/values, and the fusion representations are denoted as \( G^*(I, \hat{T}^{s}) \)
Then, a cross-lingual MLM head \(\Psi_{mlm}\), composed of a multi-layer perceptron, predicts the probability distribution \(\Psi_{mlm}(I, \hat{T}^{s})\) for the masked tokens based on the fusion representations \( G^*(I, \hat{T}^{s}) \). The source text-specific MLM pretext task is formulated as: 
\begin{equation}
\mathcal{L}^{s}_{mlm} = \mathbb{E}_{(I, T^{s}) \sim \mathcal{D}_\mathcal{M}} \mathcal{H}(y_{mlm}, \Psi_{mlm}(I, \hat{T}^{s})),
\end{equation}
where $y_{mlm}$ is a one-hot vocabulary distribution denoting the ground truth and $\mathcal{H}$ represents the cross-entropy.

For the target text \( T^{t} \), we follow the same procedure to predict its masked textual tokens based on both unmasked target textual and visual information. The target text-specific MLM pretext task is formulated as:
\begin{equation}
\mathcal{L}^{t}_{mlm} = \mathbb{E}_{(I, T^{t}) \sim \mathcal{D}_\mathcal{M}} \mathcal{H}(y_{mlm}, \Psi_{mlm}(I, \hat{T}^{t})).
\end{equation}

Taken together, the overall \emph{bi-lingual MLM} is defined as:
\begin{equation}
\mathcal{L}_{mlm} = \mathcal{L}^{s}_{mlm} + \mathcal{L}^{t}_{mlm}.
\end{equation}

Notably, in the \emph{bi-lingual MLM}, we share the same multimodal interaction encoder for both the source and target text-specific MLM computations, facilitating indirect cross-lingual relationship modeling.

\textbf{Discussion.}
There exists a structural asymmetry between the \emph{cross-lingual D-MIM} and the \emph{bi-lingual MLM}, where the distillation mechanism is integrated into the former but not the latter. 
Traditional MIM~\cite{wei2022mvp, hou2022milan}, which reconstructs images at the feature level, enables the model to better capture the semantic information of the image. 
In this process, an additional teacher model (\emph{e.g.}, CLIP) is usually introduced to extract feature representations of the image, providing supervision signals.
In our case, given the model's ability to establish robust connections between the source text and image, we utilize the multimodal interaction encoder with the input of source text and the image for supervision signals, thereby leading to \emph{cross-lingual D-MIM}. It avoids extra resource consumption and facilitates multilingual interaction during the reconstruction process, making it well-suited for our task.
In contrast, the \emph{bi-lingual MLM} performs discrete token-level reconstruction using ground-truth token IDs as strong supervision, facilitating direct learning of high-level textual semantics. Applying distillation in this context—by aligning the student's reconstructed target text with the teacher's fused source text-image representations—is fundamentally problematic. Unlike continuous visual features, textual tokens are discrete and sparsely indexed in the vocabulary. Semantically equivalent tokens across languages occupy unrelated indices, rendering the teacher's output distribution over token IDs a misaligned and incoherent signal for the student. This semantic and index-level misalignment not only undermines the distillation objective but may introduce significant label noise, making such an approach methodologically unsound.

\subsection{Multi-dimensional Global Alignment}
Beyond the Bi-IRR module focusing on fine-grained alignment learning, the Md-GA module, comprising the \emph{bi-lingual ITC} and \emph{bi-lingual A-ITM} pretext tasks, serves as a complement to bridge the modality heterogeneity between vision and language at the coarse-grained level.
The \emph{bi-lingual ITC} aims to align global representations across languages and modalities from the unimodal encoders.
The \emph{bi-lingual A-ITM} focuses on aligning cross-lingual cross-modal fusion representations from the multimodal interaction encoder.

\textbf{Bi-Lingual Image-Text Contrastive}. 
We adopt the \emph{bi-lingual ITC} to pull positive image-text samples together while pushing negative ones apart.
Specifically, considering that the \(n\)-th triple data \((I_n, T^s_n, T^t_n)\) contains two image-text pairs, \emph{i.e.}, \((I_n, T^s_n)\) and \((I_n, T^t_n)\), we conduct ITC on each pair individually.
Taking \((I_n, T^s_n)\) for example, we first obtain the global representations of \(I\) and \(T^s\) as \(v_{cls}\) and \(s_{cls}\) through the unimodal encoders, respectively. 
Based on them, we compute the image-to-text and text-to-image similarities, defined as follows:
\begin{equation} \label{eq:10}
    p^{i2t}_n = \frac{\exp(sim(I_n, T^{s}_n) / \tau)}{\sum^{N}_{j=1} \exp(sim(I_n, T^{s}_j) / \tau)},
\end{equation}
\begin{equation} \label{eq:11}
    p^{t2i}_n = \frac{\exp(sim(T^{s}_n, I_n) / \tau)}{\sum^{N}_{j=1} \exp(sim(T^{s}_n, I_j) / \tau)},
\end{equation}
where \( sim(I_n, T^{s}_n) = {h_{v}(v_{cls})}^{\top} h_{s}(s_{cls}) \), \(h_v(\cdot)\) and \(h_s(\cdot)\), implemented as two linear projection layers, project the global representations into a lower-dimensional space. \( \tau \) is a learnable temperature parameter.
The ITC pretext task on the pair \((I, T^s)\) thus is computed as:
\begin{equation} \label{eq:9}
    \mathcal{L}^{s}_{itc} = \frac{1}{2} \mathbb{E}_{(I, T^{s}) \sim \mathcal{D}_\mathcal{M}} \left[ \mathcal{H}(y^{i2t}, p^{i2t}) + \mathcal{H}(y^{t2i}, p^{t2i}) \right],
\end{equation}
where \( y^{i2t} \) and \( y^{t2i} \) are the normalized ground truth labels. 
The ITC pretext task on \((I, T^t)\) performs similarly, resulting in \(\mathcal{L}^{t}_{itc}\).

Thus, the overall \emph{bi-lingual ITC} is given by:
\begin{equation}
    \mathcal{L}_{itc} = \frac{1}{2} (\mathcal{L}^{s}_{itc} + \mathcal{L}^{t}_{itc}).
\end{equation}

\textbf{Bi-Lingual Asymmetric Image-Text Matching}. 
We also leverage \emph{bi-lingual A-ITM} to predict whether an image-text pair is matched.
Specifically, for the image-text pair \((I, T^s)\), we first obtain the global fusion representation \(g^{s}_{cls}\) from \(G(I, T^s)\).
Then, \(g^{s}_{cls}\) is fed into a multi-layer perceptron \(\Psi_{itm}\) to predict the matching probability \(\Psi_{itm} (I, T^s)\), and the ITM on \((I, T^s)\) is formulated as:
\begin{equation}
    \mathcal{L}^{s}_{itm} = \mathbb{E}_{(I,T^{s}) \sim \mathcal{D}_\mathcal{M}} \mathcal{H}(y^{itm}, \Psi_{itm} (I, T^{s})),
\label{eq: 12}
\end{equation}
where $y^{itm}$ is a two-dimensional one-hot vector representing the ground truth label.

On the other hand, for the image-text pair \((I, T^t)\), we also compute the global fusion representation \(g^{t}_{cls}\) in \(G(\hat{I}, T^t)\).
Notably, the target text representations \(F(T^s)\) and the masked image representations \(F(\hat{I})\) are employed as input to the multimodal interaction encoder. Thus, the ITM on \((I, T^t)\) is formulated as:
\begin{equation}
    \mathcal{L}^{t}_{itm} = \mathbb{E}_{(I, T^{t}) \sim \mathcal{D}_\mathcal{M}} \mathcal{H}(y^{itm}, \Psi_{itm} (\hat{I}, T^{t})).
\label{eq: 13}
\end{equation}

Overall, the \emph{bi-lingual A-ITM} pretext task is defined as:
\begin{equation}
    \mathcal{L}_{a\textit{-}itm} = \frac{1}{2} (\mathcal{L}^{s}_{itm} + \mathcal{L}^{t}_{itm}).
\end{equation}

\begin{figure*}[t]
\includegraphics[width=0.97\textwidth, height=0.44\textwidth]{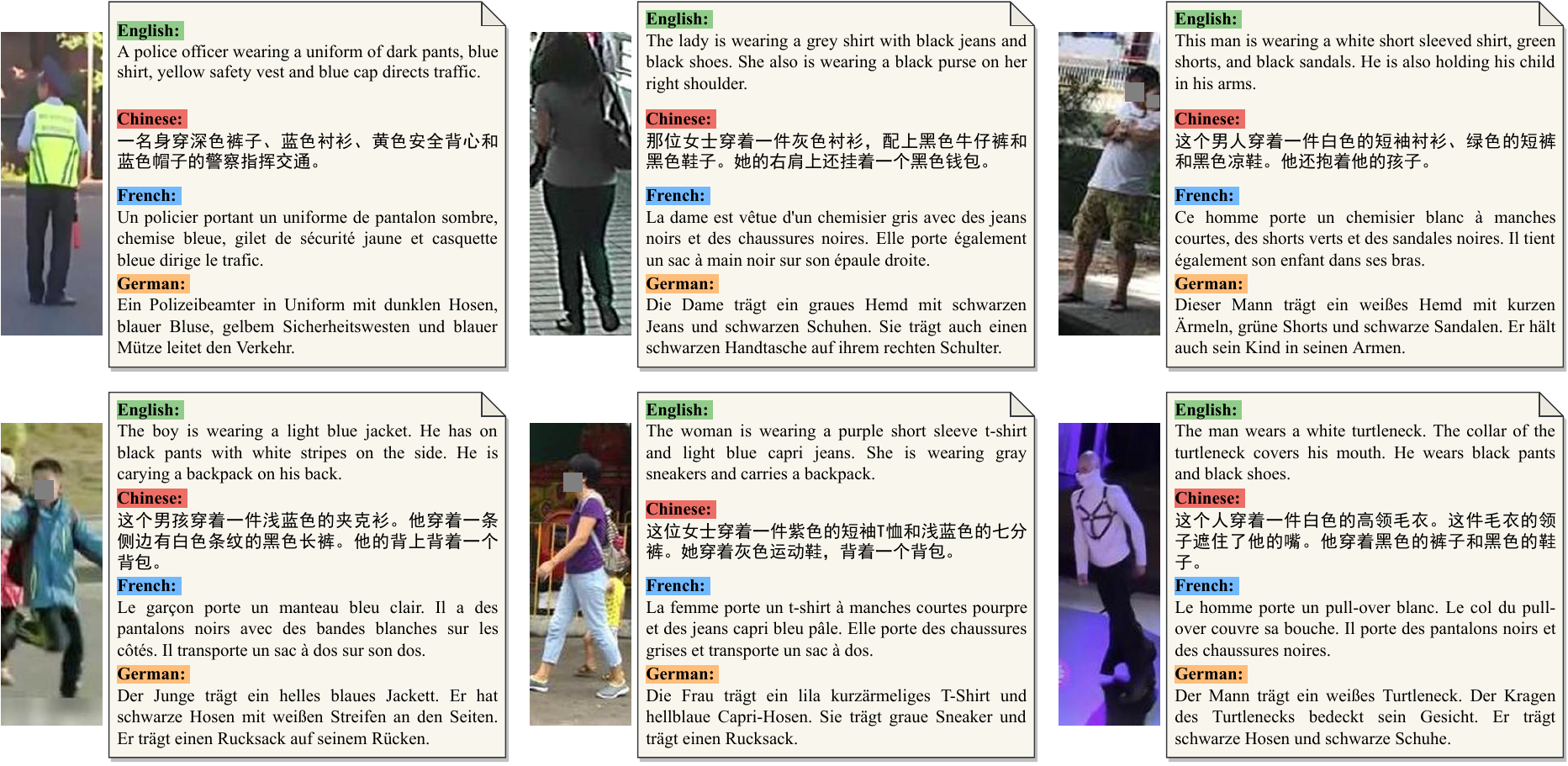}
\caption{Examples from CUHK-PEDES(M). Each image is paired with text descriptions in four different languages.}
\label{fig: multi example cuhk}
\end{figure*}

Here, an asymmetric masking design is implemented in computing the two ITM tasks. 
Specifically, the image representations paired with the target text are masked, whereas those paired with the source text are unmasked.
Despite efforts to denoise the generated target texts in LDAT, there is inevitably some noisy correspondence between the target text and image data.
By applying masking exclusively on the target text branch, \emph{i.e.,} randomly masking a subset of image tokens paired with the target text, we effectively reduce the model's reliance on potentially noisy image-text alignments. This introduces a form of semantic regularization, encouraging the model to learn robust, holistic representations by reasoning over partial visual input. In this way, masking serves as a regularizer that enhances generalization under noisy supervision.

\subsection{Joint Optimization}
Finally, we combine the Bi-IRR module with the Md-GA module and formulate the joint optimization objective:
\begin{equation}
    \mathcal{L} = \underbrace{\mathcal{L}_{itc} + \lambda_{1} \mathcal{L}_{a\textit{-}itm}}_{\text{Md-GA}} + \underbrace{\mathcal{L}_{mlm} + \lambda_{2} \mathcal{L}_{d\text{-}mim}}_{\text{Bi-IRR}},
    \label{eq: 15}
\end{equation}
where \(\lambda_1\) and \(\lambda_2\) are the hyper-parameters.



\section{Experiment}

\subsection{Datasets and Evaluation Metrics}

The current published TIPR datasets include CUHK-PEDES~\cite{li2017person}, ICFG-PEDES~\cite{ding2021semantically}, RSTPReID~\cite{zhu2021dssl}, and UFineBench~\cite{zuo2024ufinebench}, all featuring text queries exclusively in English.
Building upon these datasets, we employ the proposed LDAT to build the corresponding multilingual TIPR datasets: CUHK-PEDES(M), ICFG-PEDES(M), RSTPReID(M), and UFineBench(M). These new datasets incorporate text queries in English, Chinese, French, and German.
Notably, we conduct manual inspection and revision on the test set of these datasets to ensure evaluation quality. 
Fig.~\ref{fig: multi example cuhk} showcases some data examples from CUHK-PEDES(M).

\begin{table*}[t]\small
    \centering
    \caption{Performance comparisons on English TIPR with state-of-the-art TIPR methods on CUHK-PEDES, ICFG-PEDES, and RSTPREid. We categorize these methods into two groups based on whether they are pre-trained on large-scale person data. The method indicated by * is trained on the corresponding multilingual dataset.}
    \small{
    \resizebox{0.95\linewidth}{!}{
        \renewcommand\arraystretch{1.2}

    \begin{tabular}{l|l|cccc|cccc|cccc}
    \hline\thickhline
    \rowcolor{lightgray}
    &  & \multicolumn{4}{c|}{CUHK-PEDES} & \multicolumn{4}{c|}{ICFG-PEDES} & \multicolumn{4}{c}{RSTPReid} \\  
    \rowcolor{lightgray}
    \multirow{-2}{*}{Methods}    &   \multirow{-2}{*}{Reference}       & R@1   & R@5   & R@10   & mAP    & R@1   & R@5   & R@10   & mAP    & R@1   & R@5   & R@10  & mAP   \\ 
    \hline\hline
    \multicolumn{14}{l}{\textit{Methods without Pre-training on Large-scale Perosn Data:}} \\
    \hline\hline
    CMKA~\cite{chen2021cross}  & TIP21                      & 54.69 & 73.65 & 81.86 & -      & -     & -     & -     & - 
    & -     & -     & -    & - \\
    LapsCore~\cite{wu2021lapscore} & ICCV21                 & 63.40 & -     & 87.80 & -      & -     & -     & -     & - 
    & -     & -     & -    & - \\ 
    SAF~\cite{li2022learning}  & ICASSP22                   & 64.13 & 82.62 & 88.40 & -      & -     & -     & -     & - 
    & -     & -     & -    & - \\
    TIPCB~\cite{chen2022tipcb}  & Neuro22                   & 64.26 & 83.19 & 89.10 & -      & -     & -     & -     & - 
    & -     & -     & -    & - \\
    AXM-Net~\cite{farooq2022axm} & MM22                     & 64.44 & 80.52 & 86.77 & 58.73  & -     & -     & -     & - 
    & -     & -     & -    & - \\
    MANet~\cite{yan2023image}  & TNNLS23                    & 65.64 & 83.01 & 88.78 & -      & 59.44 & 76.80 & 82.75 & -  & -     & -     & -    & - \\
    CFine~\cite{yan2023clip}   & TIP23                      & 69.57 & 85.93 & 91.15 & -      & 60.83 & 76.55 & 82.42 & -      & 50.55 & 72.50 & 81.60 & -     \\
    IRRA~\cite{jiang2023cross} & CVPR23                     & 73.38 & 89.93 & 93.71 & 66.13  & 63.46 & 80.25 & 85.82 & 38.06  & 60.20 & 81.30 & 88.20 & 47.17  \\
    BiLMa~\cite{fujii2023bilma} & ICCV23                    & 74.03 & 89.59 & 93.62 & 66.57  & 63.83 & 80.15 & 85.74 & 38.26  & 61.20 & 81.50 & 88.80 & 48.51  \\
    RaSa~\cite{bai2023rasa}    & IJCAI23                    & 76.51 & 90.29 & 94.25 & 69.38  & 65.28 & 80.40 & 85.12 & 41.29  & 66.90 & 86.50 & 91.35 & 52.31  \\
    TBPS-CLIP~\cite{cao2024empirical}& AAAI24               & 73.54 & 88.19 & 92.35 & 65.38  & 65.05 & 80.34 & 85.47 & 39.83  & 61.95 & 83.55 & 88.75 & 48.26  \\ 
    CADA-G ~\cite{lin2024cross} & TMM24                     & 73.48 & 89.57 & 94.10 & 65.82  & 62.54 & 79.46 & 85.14 & 37.07  & 61.50 & 82.60 & 89.15 & 47.28    \\
    UMSA~\cite{zhao2024unifying}& AAAI24                    & 74.25 & 89.83 & 93.58 & 66.15  & 65.62 & 80.54 & 85.83 & 38.78  & 63.40 & 83.30 & 90.30 & 49.28  \\ 
    FSRL~\cite{wang2024fine}   & ICMR24                     & 74.86 & 89.97 & 94.14 & 67.57  & 64.93 & 80.71 & 86.19 & 40.67  & 60.65 & 83.05 & 89.60 & 48.18  \\
    Propot~\cite{yan2024prototypical} & MM24                & 74.89 & 89.90 & 94.17 & 67.12  & 65.12 & 81.57 & 86.97 & 42.93  & 61.87 & 83.63 & 89.70 & 47.82  \\
    CFAM~\cite{zuo2024ufinebench}& CVPR24                   & 75.60 & 90.53 & 94.36 & 67.27  & 65.38 & 81.17 & 86.35 & 39.42  & 62.45 & 83.55 & 91.10 & 49.50   \\
    RDE~\cite{qin2024noisy}    & CVPR24                     & 75.94 & 90.14 & 94.12 & 67.56  & 67.68 & 82.47 & 87.36 & 40.06  & 65.35 & 83.95 & 89.90 & 50.88  \\ 
    \hdashline
    IRRA*~\cite{jiang2023cross}& CVPR23                     & 64.05 & 82.91 & 88.73 & 58.44 & 57.14 & 75.37 & 82.06 & 34.46
          & 46.35 & 68.35 & 78.25 & 38.76  \\
    \hline\rowcolor[HTML]{D7F6FF}
    Bi-IRRA*                       & Ours                    & \textbf{78.82} & \textbf{92.02} & \textbf{95.47} & \textbf{69.68}  & \textbf{68.53} & \textbf{83.04} & \textbf{87.79} & \textbf{41.82}  & \textbf{72.85} & \textbf{87.75} & \textbf{91.90} & \textbf{55.60}  \\
    \hline\hline
    \multicolumn{14}{l}{\textit{Methods with Pre-training on Large-scale Perosn Data:}} \\
    \hline\hline
    PLIP~\cite{zuo2023plip}    & Arxiv23                    & 75.36 & 90.86 & 94.87  &  -     & 66.17 & 83.37 & 88.94  &   -    &  -    & -     & -     & -      \\
    APTM~\cite{yang2023towards}& MM23                       & 76.53 & 90.04 & 94.15  & 66.91  & 68.51 & 82.99 & 87.56  & 41.22  & 67.50 & 85.70 & 91.45 & 52.56  \\
    DP~\cite{song2024diverse}  & AAAI24                     & 75.66 & 90.59 & 94.07 & 66.58  & 65.61 & 81.73 & 86.95 & 39.14  & 62.48 & 83.77 & 89.93 & 48.86  \\ 
    AUL~\cite{li2024adaptive}  & AAAI24                     & 77.23 & 90.43 & 94.41 &   -    & 69.16 & 83.32 & 88.37  &   -    & 71.65 & 87.55 & 92.05 &   -    \\
    MLLM+IRRA~\cite{tan2024harnessing}& CVPR24         & 76.82 & 91.16 & 94.46 & 69.55 & 67.05 & 82.16 & 87.33 & 41.51 & 68.50 & 87.15 & 92.10 & 53.02  \\
    MLLM+APTM~\cite{tan2024harnessing}& CVPR24         & 78.13 & 91.19 & 94.50 & 68.75 & 69.37 & 83.55 & 88.18 & 42.42  & 69.95 & 87.35 & 92.30 & 54.17  \\
    \hdashline
    MLLM+IRRA*~\cite{tan2024harnessing}& CVPR24        & 68.41 & 86.21 & 91.13 & 62.33 & 60.10 & 76.88 & 83.23 & 37.04
          & 51.15 & 72.90 & 80.75 & 41.74 \\
    \hline\rowcolor[HTML]{D7F6FF}
    Bi-IRRA*      & Ours                       & \textbf{79.43} & \textbf{92.59} & \textbf{95.68} & \textbf{70.51} & \textbf{70.36} & \textbf{83.86} & \textbf{88.47} & \textbf{43.28}  & \textbf{72.50} & \textbf{88.15} & \textbf{92.45} & \textbf{57.32} \\
    \hline\thickhline
    \end{tabular}
    }}
    \label{tab: sota tipr}
\end{table*}

\begin{table}[t]\small
    \centering
    \caption{Performance comparisons on English TIPR with state-of-the-art TIPR methods on UFineBench. These results are obtained without pre-training on person data. The method indicated by * is trained on the corresponding multilingual dataset.}
    \small{
    \resizebox{0.9\linewidth}{!}{
        \renewcommand\arraystretch{1.2}
    
    \begin{tabular}{l|l|cccc}
    \hline\thickhline
    \rowcolor{lightgray}
    Methods                         & Reference & R@1    & R@5     & R@10   & mAP    \\ 
    \hline\hline
    NAFS~\cite{gao2021contextual}   & ECCV18    & 64.11  & 80.32  & 85.05  & 63.47  \\
    SSAN~\cite{ding2021semantically}& Arxiv21   & 75.09  & 88.63  & 92.84  & 73.14  \\
    LGUR~\cite{shao2022learning}    & MM22      & 70.69  & 84.57  & 89.91  & 68.93  \\
    IRRA~\cite{jiang2023cross}      & CVPR23    & 83.53  & 92.94  & 95.95  & 82.79  \\
    CFAM~\cite{zuo2024ufinebench}   & CVPR24    & 88.51  & 95.58  & 97.49  & 87.09  \\
    \hdashline
    IRRA*~\cite{jiang2023cross}              & CVPR23      & 77.20 & 90.03 & 93.92 & 76.41 \\
    \hline\rowcolor[HTML]{D7F6FF}
    Bi-IRRA*                         & Ours         & \textbf{90.45} & \textbf{96.90} & \textbf{98.18} & \textbf{89.66}  \\ 

    \hline\thickhline
    \end{tabular}
    }}
    \label{tab: ufine sota tipr}
\end{table}

\textbf{CUHK-PEDES(M)} has \(40,206\) images and \(80,440\) texts per language for \(13,003\) identities.
They are split to \(34, 054\) images and \(68, 126\) texts per language from \(11, 003\) identities in the training set, and \(3, 074\) images and \(6, 156\) texts per language from \(1, 000\) identities in the test set.

\textbf{ICFG-PEDES(M)} includes a total of \(54,522\) images for \(4,102\) identities. Each image is paired with a corresponding textual description in each language. The dataset is divided into a training set and a test set, the former comprises \(34,674\) images of \(3,102\) identities, while the latter contains \(19,848\) images for the remaining \(1,000\) identities.

\textbf{RSTPReid(M)} contains \(20,505\) images of \(4,101\) identities from \(15\) cameras. Each identity has \(5\) corresponding images taken by different cameras and each image is annotated with \(2\) textual descriptions per language. The dataset is divided into \(3,701\), \(200\), and \(200\) identities for the training, validation, and test sets, respectively.

\textbf{UFineBench(M)} contains \(26,206\) images and \(52,412\) descriptions per language of \(6,926\) identities. The dataset stands out for its ultra fine-grained textual descriptions, with text lengths \(2\)-\(3\) times longer than those of other TIPR datasets. The dataset is divided into two subsets for training and testing. The training set contains \(18,577\) images and \(37,154\) descriptions in each language. The test set contains \(7,629\) images and \(15,258\) descriptions in each language.

\textbf{Evaluation Metrics.} 
We adopt the popular Rank@$k$ (R@$k$ for short, $k=1,5,10$) to evaluate the performance of methods. In addition, for a comprehensive evaluation, we also adopt the mean Average Precision (mAP). The higher R@$k$ and mAP indicate better performance.

\subsection{Implementation Details}
\textbf{LDAT}. 
In the translation phase, Qwen~\cite{bai2023qwen} is used to translate texts from English to Chinese, while LLaMA3~\cite{dubey2024llama} handles translations from English to German and French. 
In the filtering phase, \(\theta\) is set to the mean noise level of all samples (rounded to two decimal places). In the rewriting phase, Qwen-VL~\cite{bai2023qwenvl} is utilized for English-to-Chinese translation, and Phi~\cite{abdin2024phi} is employed for translations into both German and French. These MLLMs are fine-tuned for one epoch. These specific LLMs and MLLMs are selected based on their fully open-source nature and their recognized strengths in translation accuracy and fluency.

\textbf{Bi-IRRA}. 
The parameters of the encoders used in Bi-IRRA are initialized from pre-training on a large-scale multilingual multimodal dataset~\cite{zeng2023x}. 
The text masking ratio \(p_{txt}\) in \emph{bi-lingual MLM} is configured at \(0.4\), while the image masking ratio \(p_{\text{img}}\) in \emph{cross-lingual D-MIM} is set to \(0.5\). The loss weights \(\lambda_1\) and \(\lambda_2\) in Eq.~\eqref{eq: 15} are set to \(4\) empirically. 
In the training of Bi-IRRA, all input images are resized to \(224 \times 224\) and augmented with techniques from TBPS-CLIP~\cite{cao2024empirical}. We use the AdamW optimizer with a linear warmup and cosine decay schedule, starting from an initial learning rate of \(1e-6\), decaying to \(5e-6\), and peaking at \(5e-5\). The model is trained for \(10\) epochs. For the CUHK-PEDES(M), ICFG-PEDES(M), and RSTPReid(M) datasets, we set the textual token length \(L\) to \(77\) and batch size to \(32\). For the UFineBench(M) dataset, which contains longer text sequences, we set \(L\) to \(168\), following the setting of UFineBench~\cite{zuo2024ufinebench}, and reduce the batch size to \(16\) accordingly. All experiments are conducted on four A40 GPUs.

\subsection{Comparison with State-of-the-Art Methods}
This section presents the comparative results with state-of-the-art methods on four datasets.
The proposed Bi-IRRA is the first attempt at multilingual TIPR and other methods are tailored for English TIPR exclusively. Thus we first compare Bi-IRRA with traditional TIPR methods to verify its effectiveness on English TIPR.
Subsequently, we extend the comparison to include multilingual image-text retrieval (MITR) methods to assess performance in non-English TIPR. 

\begin{table*}[t]\small
    \centering
    \caption{Performance comparisons with state-of-the-art MITR methods on CUHK-PEDES(M), ICFG-PEDES(M), and RSTPReid(M). The results of other methods are reproduced by running the official code on these multilingual TIPR datasets.}
    \small{
    \resizebox{0.98\linewidth}{!}{
        \renewcommand\arraystretch{1.2}

    \begin{tabular}{c|l|l|cccc|cccc|cccc}
    \hline\thickhline
    \rowcolor{lightgray}
    &   &   & \multicolumn{4}{c|}{CUHK-PEDES(M)} & \multicolumn{4}{c|}{ICFG-PEDES(M)} & \multicolumn{4}{c}{RSTPReid(M)} \\
    \rowcolor{lightgray}
    \multirow{-2}{*}{Language}    & \multirow{-2}{*}{Methods}    &   \multirow{-2}{*}{Reference}       & R@1   & R@5   & R@10   & mAP    & R@1   & R@5   & R@10   & mAP    & R@1   & R@5   & R@10  & mAP   \\ 
    \hline\hline
    \multirow{6}{*}{Chinese} 
        & IRRA~\cite{jiang2023cross}   & CVPR23      
            & 66.52 & 84.58 & 90.85 & 60.56 & 56.83 & 74.99 & 81.42 & 34.66 & 51.60 & 74.15 & 82.45 & 41.91 \\
        & MLLM+IRRA~\cite{tan2024harnessing}      & CVPR24      
            & 70.63 & 87.22 & 92.24 & 64.01 & 59.71 & 76.58 & 82.78 & 37.00 & 56.40 & 75.95 & 83.45 & 45.81  \\
        \cdashline{2-15} 
        & CCLM~\cite{zeng2022cross}    & ACL23   
            & 70.83 & 87.78 & 92.85 & 63.14 & 61.83 & 78.06 & 83.98 & 34.01 & 68.60 & 86.15 & 90.85 & 52.63 \\
        & X$^2$-VLM~\cite{zeng2023x}   & TPAMI23 
            & 74.81 & 90.09 & 94.17 & 66.00  & 66.54 & 81.08 & 86.28 & 39.19  & \textbf{72.00} & 86.65 & 91.35 & 54.67 \\
        & CCRK~\cite{nie2024improving} & KDD24   
            & 68.31 & 86.24 & 91.91 & 60.69 & 57.98 & 75.65 & 81.99 & 30.91 & 64.05 & 84.70 & 90.90 & 49.73  \\
    \hhline{|~|-|-|-|-|-|-|-|-|-|-|-|-|-|-|}\rowcolor[HTML]{D7F6FF}
    \cellcolor[HTML]{FFFFFF}& Bi-IRRA  &   Ours                      
        & \textbf{76.43} & \textbf{90.72} & \textbf{94.79} & \textbf{67.79}  & \textbf{66.79} & \textbf{81.55} & \textbf{86.44} & \textbf{40.26}  & 71.75 & \textbf{87.10} & \textbf{91.65} & \textbf{55.77}  \\ \hline
    \multirow{6}{*}{French} 
        & IRRA~\cite{jiang2023cross}         & CVPR23      
            & 53.04 & 75.44 & 82.70 & 49.59 & 51.21 & 70.17 & 77.39 & 29.81 & 44.95 & 68.50 & 78.35 & 37.33 \\
        & MLLM+IRRA~\cite{tan2024harnessing} & CVPR24      
            & 64.43 & 82.75 & 89.56 & 58.34 & 53.65 & 71.76 & 78.49 & 31.66 & 49.75 & 73.75 & 83.00 & 40.47 \\
        \cdashline{2-15} 
        & CCLM~\cite{zeng2022cross} &   ACL23        
            & 71.17 & 87.61 & 92.30 & 62.89 & 60.95 & 77.75 & 83.58 & 33.16 & 69.30 & 86.65 & 91.40 & 52.15  \\
        & X$^2$-VLM~\cite{zeng2023x} & TPAMI23 
            & 75.18 & 90.37 & 94.18 & 66.17  & 65.74 & 80.53 & 85.69 & 38.59  & 71.55 & 86.95 & \textbf{91.75} & 53.69   \\
        & CCRK~\cite{nie2024improving}  &   KDD24    
            & 68.29 & 85.90 & 91.36 & 60.33 & 57.57 & 75.20 & 81.65 & 30.20 & 64.80 & 85.50 & 91.25 & 49.73 \\
        \hhline{|~|-|-|-|-|-|-|-|-|-|-|-|-|-|-|}\rowcolor[HTML]{D7F6FF}
        \cellcolor[HTML]{FFFFFF}& Bi-IRRA            &   Ours                      
            & \textbf{76.46} & \textbf{90.45} & \textbf{94.35} & \textbf{67.26}  & \textbf{66.90} & \textbf{81.48} & \textbf{86.40} & \textbf{39.44}  & \textbf{71.80} & \textbf{87.20} & 91.65 & \textbf{54.25} \\ \hline
    \multirow{6}{*}{German} 
        & IRRA~\cite{jiang2023cross}         & CVPR23      
            & 50.00 & 72.04 & 79.87 & 46.46 & 44.01 & 63.70 & 72.16 & 25.15 & 34.15 & 58.05 & 70.20 & 29.37 \\
        & MLLM+IRRA~\cite{tan2024harnessing} & CVPR24      
            & 55.07 & 75.75 & 83.61 & 50.58 & 47.27 & 66.57 & 74.12 & 27.51 & 41.30 & 65.10 & 75.20 & 34.70 \\
        \cdashline{2-15}
        & CCLM~\cite{zeng2022cross} &   ACL23        
            & 70.19 & 87.54 & 92.24 & 62.47 & 61.42 & 77.88 & 83.72 & 33.84 & 66.70 & 84.80 & 90.95 & 51.75 \\
        & X$^2$-VLM~\cite{zeng2023x} & TPAMI23 
            & 74.71 & 89.83 & 94.30 & 65.73  & 66.34 & 80.65 & 86.02 & 38.77 & 70.25 & 86.40 & 91.65 & 53.56  \\
        & CCRK~\cite{nie2024improving} &   KDD24     
            & 67.01 & 85.32 & 90.74 & 59.54 & 56.52 & 74.31 & 80.63 & 29.28 & 64.45 & 84.55 & 91.00 & 49.30 \\ 
        \hhline{|~|-|-|-|-|-|-|-|-|-|-|-|-|-|-|}\rowcolor[HTML]{D7F6FF}
        \cellcolor[HTML]{FFFFFF} & Bi-IRRA            &   Ours                      
            & \textbf{75.57} & \textbf{90.56} & \textbf{94.31} & \textbf{67.07}  & \textbf{67.05} & \textbf{81.68} & \textbf{86.62} & \textbf{39.91}  & \textbf{70.95} & \textbf{87.45} & \textbf{92.10} & \textbf{54.92} \\
    
    \hline\thickhline
    \end{tabular}
    }}
    \label{tab: sota mitr}
\end{table*}
\begin{table}[t]\small
    \centering
    \caption{Performance comparisons on non-English TIPR with state-of-the-art MITR methods on UFineBench(M).}
    \small{
    \resizebox{0.98\linewidth}{!}{
        \renewcommand\arraystretch{1.2}

    \begin{tabular}{c|l|l|cccc}
    \hline\thickhline
    \rowcolor{lightgray}
    Language                      & Methods & Reference & R@1 & R@5 & R@10 & mAP \\
    \hline\hline

    \multirow{6}{*}{Chinese} 
        & IRRA~\cite{jiang2023cross}         & CVPR23   & 79.01 & 90.96 & 94.66 & 78.24 \\
        \cdashline{2-7}
        & CCLM~\cite{zeng2022cross}          & ACL23    & 84.98 & 94.03 & 96.44 & 83.37 \\
        & X$^2$-VLM~\cite{zeng2023x}         & TPAMI23  & 88.20 & 96.03 & 97.90 & 87.12 \\
        & CCRK~\cite{nie2024improving}       & KDD24    & 79.58 & 91.12 & 94.49 & 77.98 \\
        \hhline{|~|-|-|-|-|-|-|}\rowcolor[HTML]{D7F6FF}
        \cellcolor[HTML]{FFFFFF}
        & Bi-IRRA  &   Ours       &  \textbf{89.98} & \textbf{96.56} & \textbf{98.05} & \textbf{89.22} \\ \hline
    \multirow{6}{*}{French} 
        & IRRA~\cite{jiang2023cross}         & CVPR23   & 70.32 & 85.05 & 90.47 & 69.80 \\
        \cdashline{2-7}
        & CCLM~\cite{zeng2022cross}          & ACL23    & 83.58 & 93.22 & 95.84 & 81.80 \\
        & X$^2$-VLM~\cite{zeng2023x}         & TPAMI23  & 87.58 & 95.45 & 97.39 & 86.21 \\
        & CCRK~\cite{nie2024improving}       & KDD24    & 78.61 & 90.48 & 94.20 & 76.84 \\
        \hhline{|~|-|-|-|-|-|-|}\rowcolor[HTML]{D7F6FF}
        \cellcolor[HTML]{FFFFFF}
        & Bi-IRRA  &   Ours       & \textbf{89.30} & \textbf{96.24} & \textbf{97.73} & \textbf{88.32}  \\ \hline
    \multirow{6}{*}{German} 
        & IRRA~\cite{jiang2023cross}         & CVPR23   & 59.40 & 77.03 & 84.02 & 59.63 \\
        \cdashline{2-7}
        & CCLM~\cite{zeng2022cross}          & ACL23    & 83.42 & 93.18 & 95.77 & 81.49  \\
        & X$^2$-VLM~\cite{zeng2023x}         & TPAMI23  & 86.86 & 95.31 & 97.29 & 85.52  \\
        & CCRK~\cite{nie2024improving}       & KDD24    & 77.63 & 90.33 & 93.98 & 75.87  \\
        \hhline{|~|-|-|-|-|-|-|}
        \rowcolor[HTML]{D7F6FF}
        \cellcolor[HTML]{FFFFFF}
        & Bi-IRRA  &   Ours        & \textbf{89.39} & \textbf{96.07} & \textbf{97.75} & \textbf{88.02}  \\
    
    \hline\thickhline
    \end{tabular}
    }}
    \label{tab: ufine sota mitr}
\end{table}

\textbf{Performance Comparison on English TIPR}. 
We report the results on CUHK-PEDES, ICFG-PEDES and RSTPReid in Table~\ref{tab: sota tipr}. Considering that some TIPR methods improve performance by pre-training on large-scale person data (\emph{e.g.}, MALS~\cite{yang2023towards} and LUPerson-MLLM~\cite{tan2024harnessing}), we conduct comparisons under both scenarios: with and without pre-training.

Whether compared to methods with or without pre-training, Bi-IRRA achieves the highest performance.
In particular, Bi-IRRA pretrained on LUPerson-MLLM~\cite{tan2024harnessing} surpasses the current state-of-the-art method, MLLM+APTM~\cite{tan2024harnessing} by \(1.30\)\%/\(1.76\)\%, \(0.99\)\%/\(0.86\)\%, and \(2.55\)\%/\(3.15\)\% in terms of R@1/mAP on three datasets, respectively. 
Compared with the state-of-the-art method RaSa~\cite{bai2023rasa}, Bi-IRRA without pre-training gains a significant R@1 improvement of \(2.31\%\), \(3.25\%\), and \(5.95\%\) on the three datasets, respectively.
It also outperforms the previous conference work IRRA~\cite{jiang2023cross} by a significant margin.
These results highlight the superiority of the proposed Bi-IRRA.
Bi-IRRA is specifically tailored to be adaptive for multilingual TIPR data, and training it on these multilingual datasets enhances its cross-modal English-image modeling capabilities to some extent. In contrast, traditional TIPR methods are exclusively trained on English data.
We also reproduce the results of several TIPR methods\footnote{To adapt these traditional TIPR methods for multilingual data during training, we need to make adjustments to the tokenizer in their published code.} (IRRA~\cite{jiang2023cross}, MLLM+IRRA~\cite{tan2024harnessing}) trained on the multilingual data (indicated by * in Table~\ref{tab: sota tipr}).
Despite this, Bi-IRRA continues to demonstrate performance advantages.
These TIPR methods lack the model architecture designed for effectively learning from multilingual data, resulting in subpar results even when trained on such datasets.

Moreover, we present the results on UFineBench which involves longer texts with ultra-fine-grained information in Table~\ref{tab: ufine sota tipr}. Bi-IRRA still showcases superior performance on this dataset. Specifically, Bi-IRRA outperforms the current state-of-the-art method CFAM~\cite{zuo2024ufinebench} by the significant margins of \(1.94\)\%/\(2.57\)\% at R@1/mAP.

\textbf{Performance Comparison on non-English TIPR}. 
We compare Bi-IRRA with other methods on non-English TIPR, including some TIPR methods trained on multilingual corpora (IRRA~\cite{jiang2023cross}, MLLM+IRRA~\cite{tan2024harnessing}) and state-of-the-art MITR methods (CCRK~\cite{nie2024improving}, CCLM~\cite{zeng2022cross} and X$^2$-VLM~\cite{zeng2023x}). The results are shown in Tables~\ref{tab: sota mitr} and \ref{tab: ufine sota mitr}.  
Bi-IRRA consistently shows superior performance across various non-English environments. 
For instance, on the CUHK-PEDES(M) dataset, the Bi-IRRA method surpasses the state-of-the-art MITR method X\(^2\)-VLM~\cite{zeng2023x} by \(1.62\%\)/\(1.79\%\), \(1.28\%\)/\(1.09\%\), and \(0.86\%\)/\(1.34\%\) in terms of R@1/mAP in Chinese, French, and German, respectively.
It indicates that Bi-IRRA achieves effective alignment between different languages and images, demonstrating strong robustness and generalization.

\begin{table}[t]\small
    \caption{Ablation study on the proposed LDAT on CUHK-PEDES(M). Trans. is the abbreviation of translation.}
    \small{
    \resizebox{1.0\linewidth}{!}{
        \renewcommand\arraystretch{1.2}
    
    \begin{tabular}{l|cccc|cccc}
    \hline\thickhline
    \rowcolor{lightgray}
    &      \multicolumn{4}{c|}{English} & \multicolumn{4}{c}{Chinese}  \\
    \rowcolor{lightgray}
    \multirow{-2}{*}{}         & R@1 & R@5 & R@10 & mAP & R@1 & R@5 & R@10 & mAP    \\ 
    \hline\hline
    LLMs-driven Trans.         & 78.09 & 91.94 & 95.52 & 69.37 & 75.89 & 90.58 & 94.57 & 67.53 \\
    MLLMs-driven Trans.        & 78.61 & 92.02 & 95.47 & 69.55 & 76.19 & 90.85 & 94.61 & 67.78 \\
    LDAT (w/ LLM)              & 78.22 & 91.96 & 95.42 & 69.27 & 76.02 & 90.46 & 94.54 & 67.64  \\
    \hline
    \rowcolor[HTML]{D7F6FF}
    LDAT                       & \textbf{78.82} & \textbf{92.02} & \textbf{95.47} & \textbf{69.68} & \textbf{76.43} & \textbf{90.72} & \textbf{94.79} & \textbf{67.79} \\

    \hline\thickhline
    \end{tabular}
    }}
    \label{tab: ldat components}
\end{table}

\begin{table*}[b]\small
    \centering
    \caption{Ablation study on components of Bi-IRRA. Bi-IRRA consists of two main modules: Bi-IRR and Md-GA. The Bi-IRR module is composed of bi-lingual MLM and cross-lingual D-MIM. The Similarity Distribution Matching (SDM) and ID pretext tasks are employed in the previous conference version, similar to the Md-GA module in this work, both of which focus on cross-modal global alignment.}
    \small{
    \resizebox{0.95\linewidth}{!}{
        \renewcommand\arraystretch{1.2}

    \begin{tabular}{c|cc:ccc|cccc|cccc}
    \hline\thickhline
    \rowcolor{lightgray}
    & \multicolumn{2}{c:}{Bi-IRR}  & & & & \multicolumn{4}{c|}{English} & \multicolumn{4}{c}{Chinese}  \\
    \rowcolor{lightgray}
    \multirow{-2}{*}{No.}    & bi-lingual MLM & cross-lingual D-MIM    &  \multirow{-2}{*}{Md-GA} & \multirow{-2}{*}{SDM} & \multirow{-2}{*}{ID} & R@1   & R@5   & R@10   & mAP    & R@1   & R@5   & R@10   & mAP    \\ 
    \hline\hline
        1            &            &            & \checkmark &            &      
            & 76.97 & 91.26 & 94.96 & 67.86 & 74.81 & 90.09 & 94.17 & 66.00   \\
        2            &            & \checkmark & \checkmark &            &      
            & 77.81 & 91.83 & 95.31 & 69.16 & 75.18 & 90.29 & 94.15 & 67.08 \\
        3            & \checkmark &            & \checkmark &            &      
            & 77.49 & 91.80 & 95.14 & 68.50 & 75.36 & 90.61 & 94.49 & 66.72  \\
        \hline
        4            & \checkmark & \checkmark &            & \checkmark &       
            & 66.89 &  84.76 & 90.51 & 60.57    & 64.36 & 83.09 & 89.12 & 58.42  \\ 
        5            & \checkmark & \checkmark & \checkmark &            & \checkmark  
            & 78.28 & 91.96 & \textbf{95.50} & 69.55 & 76.20 & 90.71 & 94.62 & 67.75   \\ 
        \hline\rowcolor[HTML]{D7F6FF}
        6            & \checkmark & \checkmark & \checkmark &            &        
            & \textbf{78.82} & \textbf{92.02} & 95.47 & \textbf{69.68} & \textbf{76.43} & \textbf{90.72} & \textbf{94.79} & \textbf{67.79}  \\ 
    
    \hline\thickhline
    \end{tabular}
    }}
    \label{tab: ablation_study}
\end{table*}

\subsection{Ablation Study}
We conduct ablation studies on the proposed LDAT and Bi-IRRA to validate their effectiveness. The ablation studies are performed on CUHK-PEDES(M), with English text as the source text and Chinese text as the target text.

\begin{table*}[t]\small
    \centering
    \caption{Analysis of cross-lingual D-MIM on CUHK-PEDES(M). The cross-lingual D-MIM employs the multimodal interaction encoder with different inputs as both the teacher and student models via distilling the fusion feature representations. Additionally, the attention map generated by the multimodal interaction encoder and the CLS token of the fusion representations can act as alternatives to the fusion feature representations for distillation.}
    \small{
    \resizebox{0.9\linewidth}{!}{
        \renewcommand\arraystretch{1.2}

    \begin{tabular}{c|c|cc:cc|cccc|cccc}
    \hline\thickhline
    \rowcolor{lightgray}
    &   & \multicolumn{2}{c:}{Teacher} & \multicolumn{2}{c|}{Student}  & \multicolumn{4}{c|}{English}    &    \multicolumn{4}{c}{Chinese}     \\ 
    \rowcolor{lightgray}
    \multirow{-2}{*}{No.}    & \multirow{-2}{*}{Distill}   &  ($T^s$, $I$) & ($T^t$, $I$) &  ($T^s$, $\hat{I}$) & ($T^t$, $\hat{I}$)      & R@1   & R@5   & R@10   & mAP    & R@1   & R@5   & R@10   & mAP    \\ 
    \hline\hline
    1 & Attention Map & \checkmark &            &            & \checkmark 
    & 78.72 & 92.11 & 95.58 & 69.60 & 76.33 & 90.87 & 94.70 & 67.62 \\
    2 & CLS Token     & \checkmark &            &            & \checkmark 
    & 78.59 & 92.14 & \textbf{95.66} & 69.56 & 75.86 & 90.61 & 94.56 & 67.71 \\
    \hline
    3 & Feature       &            & \checkmark & \checkmark &            
    & 78.59 & 92.09 & 95.52 & 69.55 & 76.01 & 90.71 & 94.59 & 67.65 \\
    4 & Feature       & \checkmark &            & \checkmark &            
    & 78.38 & \textbf{92.33} & 95.58 & 69.51 & 75.97 & \textbf{90.92} & 94.64 & 67.63 \\
    5 & Feature       &            & \checkmark &            & \checkmark 
    & 78.23 & 92.14 & \textbf{95.66} & 69.57 & 75.65 & 90.79 & 94.48 & 67.61 \\
    \hdashline
    \multirow{2}{*}{6} & \multirow{2}{*}{Feature}      &  \checkmark     &  &   \checkmark      & 
    & \multirow{2}{*}{78.41} & \multirow{2}{*}{92.24} & \multirow{2}{*}{95.47} & \multirow{2}{*}{69.67} & \multirow{2}{*}{75.83} & \multirow{2}{*}{90.90} & \multirow{2}{*}{94.51} & \multirow{2}{*}{67.34} \\
      &               &            & \checkmark &            & \checkmark 
    & & & & & &  \\
    \hline
    \rowcolor[HTML]{D7F6FF}
    7 & Feature      & \checkmark &             &           & \checkmark 
    & \textbf{78.82} & 92.02 & 95.47 & \textbf{69.68} & \textbf{76.43} & 90.72 & \textbf{94.79} & \textbf{67.79} \\
    
    \hline\thickhline
    \end{tabular}
    }}
    \label{tab: sdmim}
\end{table*}

\textbf{Effectiveness of LDAT}.
We propose the LDAT pipeline, comprising translation, filtering, and rewriting phases, to automatically generate multilingual TIPR data.
To assess the effectiveness of LDAT, we experiment with three variants of automatically generating multilingual TIPR data.
In the first variant, we exclude the filtering and rewriting stages and employ LLM~\cite{dubey2024llama} for direct translation (\emph{i.e.}, without leveraging domain-specific knowledge and visual information).
In the second variant, we similarly exclude filtering and rewriting but instead use MLLM~\cite{bai2023qwenvl} for translation.
In the third variant, we replace the MLLM in the rewriting phase with the LLM~\cite{bai2023qwen, dubey2024llama}, utilizing only text information as partial domain-specific knowledge to improve translation quality.
The comparison results are presented in Table~\ref{tab: ldat components}. 
Bi-IRRA, trained on multilingual data directly generated by either LLM (LLMs-driven Trans.) or MLLM (MLLMs-driven Trans.), achieves competitive performance. That is because the textual descriptions in TIPR follow a consistent structure and the strong LLM/MLLM achieves relatively accurate translations. Notably, MLLMs-driven Trans. outperforms LLMs-driven Trans., showing the utility of visual information in translation generation.
Further improvements are observed when partial domain-specific knowledge is incorporated via LDAT (w/ LLM).
Nevertheless, the incorporation of comprehensive domain-specific knowledge by LDAT results in the best performance for Bi-IRRA.


\textbf{Ablation on Bi-IRR}. 
Bi-IRR consists of \emph{bi-lingual MLM} and \emph{cross-lingual D-MIM} pretext tasks, designed to achieve implicit local relations reasoning across languages and modalities.
We evaluate the effectiveness of these pretext tasks in Table~\ref{tab: ablation_study}.
(1) Removing all \emph{bi-lingual MLM} and \emph{cross-lingual D-MIM} pretext tasks (No.1 vs. No.6) results in a performance drop of \(1.85\%\)/\(1.82\%\) and \(1.62\%\)/\(1.79\%\) on R@1/mAP for English and Chinese, respectively.
Bi-IRR with \emph{bi-lingual MLM} and \emph{cross-lingual D-MIM} effectively captures fine-grained relations across different languages and modalities, aiding global matching in achieving better cross-modal alignment.
The exclusion of Bi-IRR can result in a lack of fine-grained relation modeling, thereby impacting performance.
(2) Removing either \emph{bi-lingual MLM} (No.2 vs. No.6) or \emph{cross-lingual D-MIM} (No.3 vs. No.6) leads to a similar decline in performance. 
The \emph{bi-lingual MLM} focuses on establishing fine-grained relations by reconstructing masked text data, while \emph{cross-lingual D-MIM} does the same for masked image data. 
Together, they form a bidirectional implicit relation reasoning that enhances the modeling of local relations between vision and language. 
Removing either of them results in partial relation reasoning and impacts performance accordingly.

\textbf{Ablation on Md-GA}. 
The Md-GA module comprises \emph{bi-lingual ITC} and \emph{bi-lingual A-ITM}, aligning global textual and visual feature representations. 
In the conference version of this work~\cite{jiang2023cross}, Similarity Distribution Matching (SDM) is employed for global alignment. Here, we replace the SDM with the Md-GA module. As shown in Table~\ref{tab: ablation_study}, it results in a significant performance improvement (No.4 vs. No.6). 
This is primarily due to the addition of \emph{bi-lingual A-ITM}, which enables the use of the fusion representations to compute similarity scores during inference. This approach captures cross-modal fine-grained information compared to directly calculating the similarity between image and text feature representations, as in the conference version work. 
Furthermore, the conference version employs ID loss~\cite{zheng2020dual} to assist unimodal encoders in learning more discriminative feature representations for enhanced global alignment. However, in this work, using ID loss (No.5 vs. No.6) does not improve the main metrics, so we omit it.

\begin{figure}[t]
\includegraphics[width=0.48\textwidth]{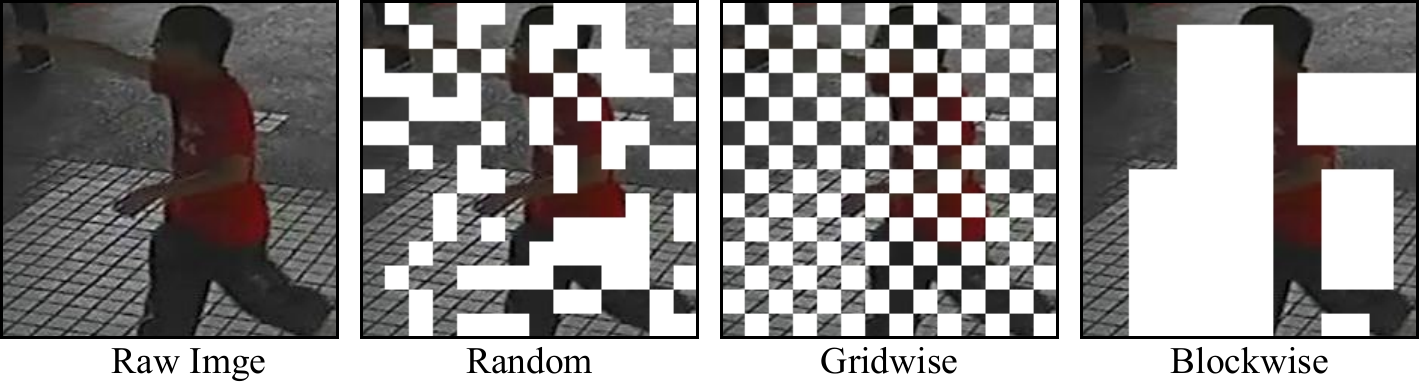}
\caption{Visualize of different masking strategies.}
\label{fig: mask_strategy}
\end{figure}

\textbf{Analysis of Cross-Lingual D-MIM}
In the Bi-IRR module, \emph{cross-lingual D-MIM} is introduced by a cross-lingual distillation mechanism, where the fusion feature representations of source text and image act as the teacher supervision to guide the reconstruction of masked images with target text.
To validate the effectiveness of the proposed \emph{cross-lingual D-MIM}, we construct several variants for comparison. The results are presented in Table~\ref{tab: sdmim}.

Rather than using feature representation for distillation, alternatives such as the cross-attention map produced by the multimodal interaction encoder (No.1) and the CLS token from the fusion feature representations (No.2) can be employed. 
However, the original \emph{cross-lingual D-MIM} using the feature for distillation (No.7) yields the best performance. This superiority can be attributed to the limited information available in the cross-attention map and the CLS token, making them insufficient for effective supervision.

In addition to the default setup of using the teacher model with \((T^s, I)\) input and the student model with \((T^t, I_M)\) input for the multimodal interaction encoder in \emph{cross-lingual D-MIM}, there exist other setups that can be explored.
(1) One such variation involves interchanging the teacher model with the student model. 
That is, the fusion feature representations of target text and image guide the reconstruction of masked images with source text (No.3). 
The results in Table~\ref{tab: sdmim} indicate a slight performance drop via this interchange operation (No.3 vs. No.6).
Given that the source text typically exhibits higher quality than the target text, the multimodal interaction encoder with \((T^s, I)\) input provides more valuable information than the encoder with \((T^t, I)\) input. Therefore, using the former as the teacher model results in better performance.
(2) Another variation involves utilizing the single text domain for \emph{cross-lingual D-MIM}. 
Specifically, we utilize the fusion representations of either the source text and image (No.4) or the target text and image (No.5) as teacher supervision to guide the reconstruction of the masked image with the corresponding text. However, both variations lead to performance decreases as they lack interactions between multiple languages.
(3) Furthermore, we can combine \emph{cross-lingual D-MIM} task with only the source text domain and that with only the target text domain, \emph{i.e.,} the sum of two separate \emph{cross-lingual D-MIM} pretext tasks (No.6).
This variation still fails to establish interactions between different languages, ultimately degrading the model's performance.

\begin{figure}[t]
\centering
\includegraphics[width=0.48\textwidth]{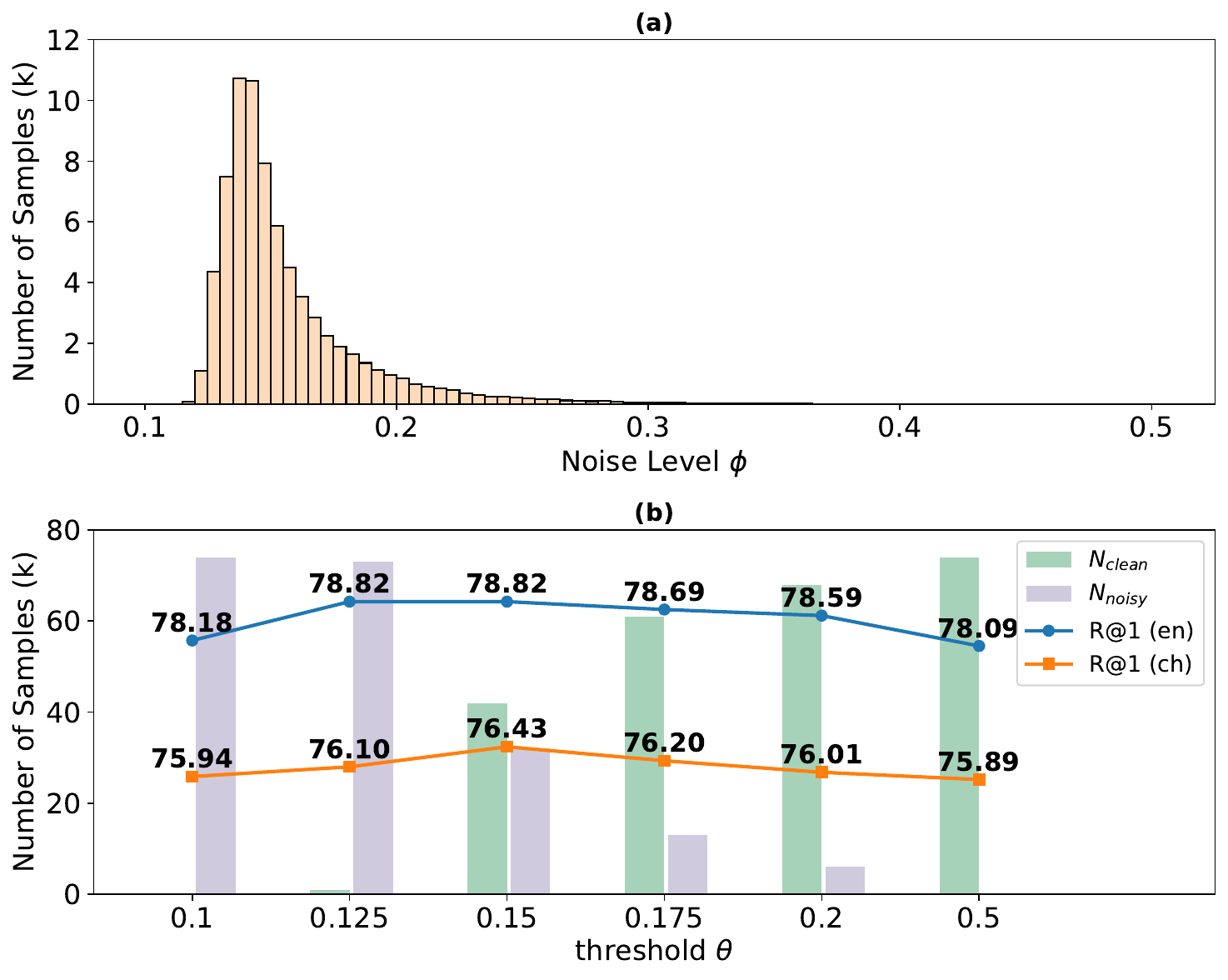}
\caption{Illustration of the distribution of noise level (a) and experimental results with varying thresholds (b). In (a), as no data falls within the noise level range of \(0.5 \sim 1\), so this interval is omitted. In (b), the green and purple bars represent the sample sizes \(N_{clean}\) and \(N_{noisy}\) of \(\mathcal{D}^C_{\mathcal{M}}\) and \(\mathcal{D}^N_{\mathcal{M}}\), respectively. The blue and orange line charts illustrate the retrieval performance in English and Chinese at the corresponding \(\theta\).}
\label{fig: dist_theta}
\end{figure}

\begin{table}[t]\small
    \centering
    \caption{Comparison of different masking strategies of SD-MIM on CUHK-PEDES(M).}
    \small{
    \resizebox{1.0\linewidth}{!}{
        \renewcommand\arraystretch{1.2}
    
    \begin{tabular}{l|cccc|cccc}
    \hline\thickhline
    \rowcolor{lightgray}
    &      \multicolumn{4}{c|}{English} & \multicolumn{4}{c}{Chinese}  \\
    \rowcolor{lightgray}
    \multirow{-2}{*}{} & R@1 & R@5  & R@10   & mAP    & R@1  & R@5   & R@10   & mAP    \\ 
    \hline\hline
    Random                             & 78.17 & 91.99 & 95.29 & 69.24 & 75.80 & \textbf{90.84} & 94.43 & 67.57 \\
    Gridwise                           & 77.89 & 91.81 & 95.14 & 68.87 & 75.68 & 90.46 & 94.69 & 67.06  \\
    \rowcolor[HTML]{D7F6FF}
    Blockwise                          & \textbf{78.82} & \textbf{92.02} & \textbf{95.47} & \textbf{69.68} & \textbf{76.43} & 90.72 & \textbf{94.79} & \textbf{67.79} \\

    \hline\thickhline
    \end{tabular}
    }}
    \label{tab: maskingStrategy}
\end{table}

Rather than utilizing the blockwise masking strategy in \emph{cross-lingual D-MIM}, alternative masking strategies like random masking and gridwise masking can also be employed. Fig.~\ref{fig: mask_strategy} illustrates these masking strategies for clarity, and Table~\ref{tab: maskingStrategy} presents the comparison results. Notably, the blockwise masking strategy yields the best results.
From Fig.~\ref{fig: mask_strategy}, the blockwise masking strategy tends to mask spatially connected image patches, increasing the likelihood of obscuring complete semantic information and making reconstruction more challenging. This encourages the model better to understand the textual information and the available visual content, thereby improving the performance.

\textbf{Analysis of Bi-Lingual A-ITM}.
In the Md-GA module, \emph{bi-lingual A-ITM} is used to constrain the global representations of fusion representations. We employ the input setting, namely \((I, T^s)\) and \((\hat{I}, T^t)\), to compute \emph{bi-lingual A-ITM}. 
Besides the default setting, we explore alternative input settings, with results reported in Table~\ref{tab: itm}. 
(1) We first employ a conventional setup where \((I, T^s)\) and \((I, T^t)\) (without any mask operation) are used to compute ITM (No.1), and yet resulting in limited performance.
(2) We experiment with masking input image data during ITM computation. As shown in No.2 \(\sim\) No.4, only masking the image paired with the target text yields the most favorable results. It facilitates noise-robust learning for the potentially noisy correspondence between the target text and image.
(3) Alternatively, we explore masking input text data during ITM computation. However, as shown in No.5 \(\sim\) No.7, the performance of masking text is generally inferior to that of masking image. The image typically exhibits more semantic coherence than the text and retains more semantic information after masking for effective cross-modal alignment.

\textbf{Analysis of Bi-Lingual ITC}.
Inspired by the masking strategy in bilingual A-ITM, we evaluate its effect on bilingual ITC. As shown in Table~\ref{tab: itc}, applying input masking in bi-lingual ITC does not yield performance improvements. While masking is known to act as a regularizer by mitigating overfitting to potentially noisy cross-modal alignments, its ineffectiveness in this context can be attributed to the architectural differences between bi-lingual ITC and bilingual A-ITM. Specifically, bi-lingual ITC computes the contrastive loss directly from encoded unimodal features, without passing them through a cross-modal interaction module. Consequently, the masked inputs do not engage in meaningful cross-modal reasoning, limiting the regularization effect.

\begin{figure}[htbp]
\centering
\includegraphics[width=0.47\textwidth]{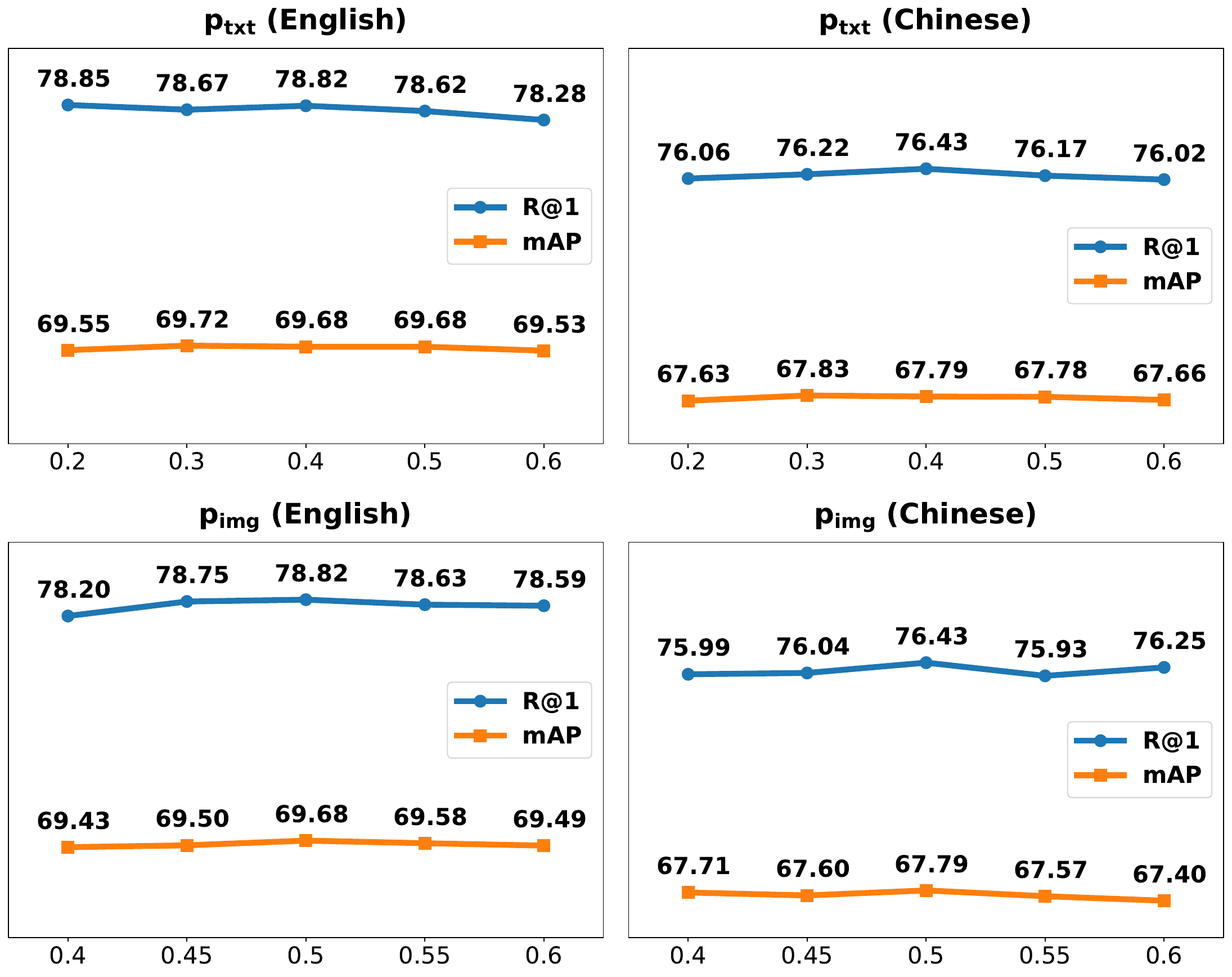}
\caption{Hyper-parameters analysis of text mask ratio \(p_{\text{txt}}\) and image mask ration \(p_{\text{img}}\).}
\label{fig: mask ratio}
\end{figure}

\begin{figure}[htbp]
\centering
\includegraphics[width=0.47\textwidth]{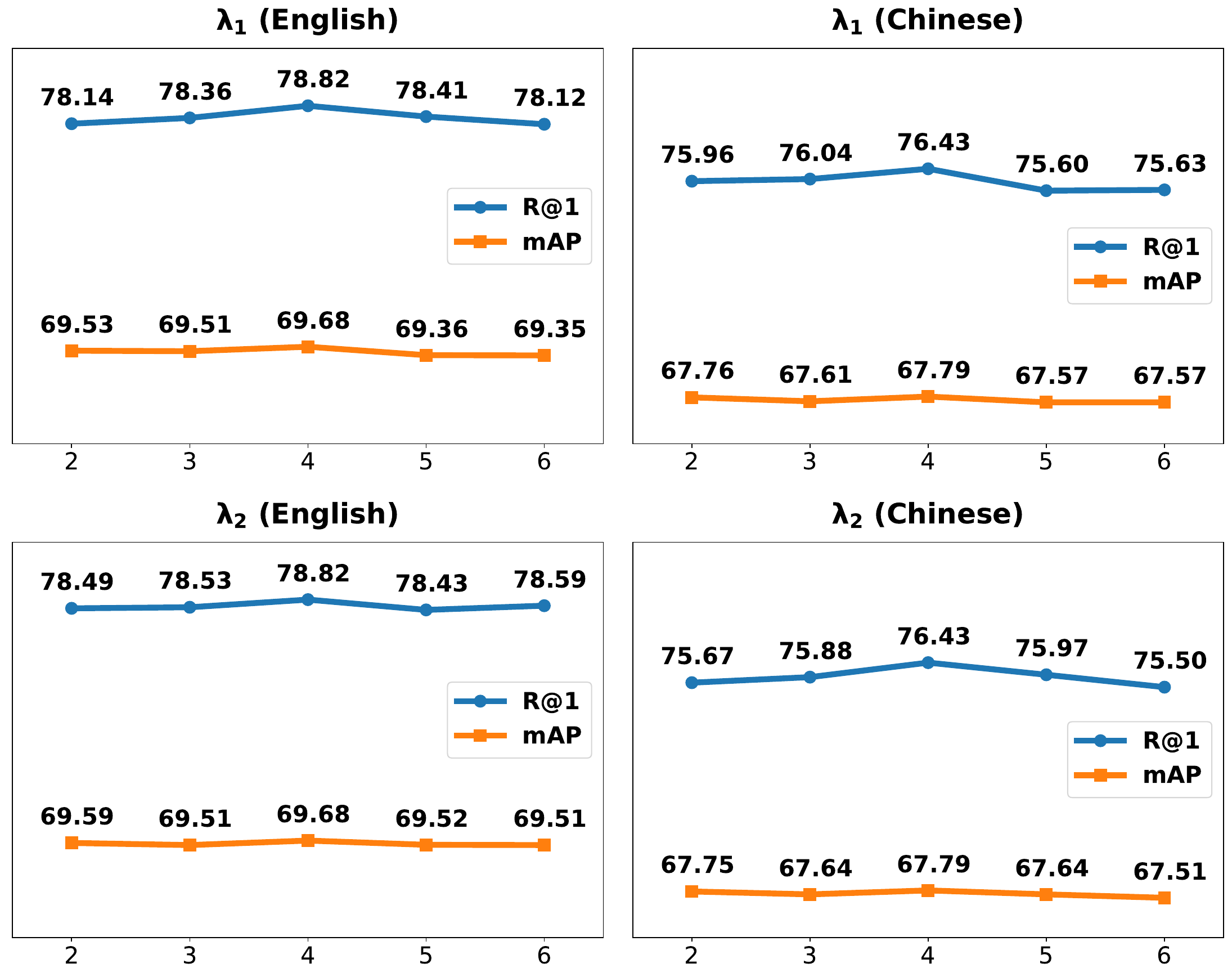}
\caption{Hyper-parameters analysis of loss weights \(\lambda_1\) and \(\lambda_2\) on CUHK-PEDES(M).}
\label{fig: hype}
\end{figure}

\begin{table}[t]\small
    \centering
    \caption{Analysis of bi-lingual A-ITM on CUHK-PEDES(M). Input-1 and Input-2 are the inputs for Eq.~\eqref{eq: 12} and Eq.~\eqref{eq: 13}, respectively. A special mark \([M]\) is used to indicate that \(T^s\), \(T^t\), and \(I\) are masked during computation.}
    \small{
    \resizebox{0.99\linewidth}{!}{
        \renewcommand\arraystretch{1.2}

    \begin{tabular}{c|cc:cc|ccc|ccc}
    \hline\thickhline
    \rowcolor{lightgray}
    &   \multicolumn{2}{c:}{Input-1} & \multicolumn{2}{c|}{Input-2}   & \multicolumn{3}{c|}{English}    &    \multicolumn{3}{c}{Chinese}     \\ 
    \rowcolor{lightgray}
    \multirow{-2}{*}{No.} & $I$ & $T^s$ & $I$ & $T^t$
        & R@1  & R@10  & mAP  & R@1 & R@10  & mAP    \\ 
    \hline\hline
    1   &     &     &     & 
        & 77.92 & 95.24 & 68.71 & 75.49 & 94.43 & 66.83 \\
    \hline
    2   & [M] &    &     &                            
        & 78.53 & \textbf{95.61} & 69.60  & 76.17 & 94.69 & 67.59  \\
    \rowcolor[HTML]{D7F6FF}
    3   &     &    & [M] & 
        & \textbf{78.82} & 95.47 & \textbf{69.68} & \textbf{76.43} & \textbf{94.79} & \textbf{67.79} \\
    4   & [M] &    & [M] & 
        & 77.55 & 95.47 & 69.38 & 75.32 & 94.41 & 67.52 \\
    \hline
    5   &     & [M] &     &                            
        & 76.95 & 95.11 & 67.86 & 75.91 & 94.57 & 66.98  \\
    6   &     &     &     & [M]
        & 78.14 & 95.22 & 68.63 & 75.13 & 94.22 & 66.42 \\
    7   &     & [M] &     & [M]
        & 73.86 & 94.04 & 65.54 & 72.55 & 93.29 & 64.67 \\
    
    \hline\thickhline
    \end{tabular}
    }}
    \label{tab: itm}
\end{table}

\begin{table}[ht]\small
    \centering
    \caption{Analysis of bi-lingual ITC on CUHK-PEDES(M).}
    \small{
    \resizebox{0.99\linewidth}{!}{
        \renewcommand\arraystretch{1.2}

    \begin{tabular}{c|cc:cc|ccc|ccc}
    \hline\thickhline
    \rowcolor{lightgray}
    &   \multicolumn{2}{c:}{Input-1} & \multicolumn{2}{c|}{Input-2}   & \multicolumn{3}{c|}{English}    &    \multicolumn{3}{c}{Chinese}     \\ 
    \rowcolor{lightgray}
    \multirow{-2}{*}{No.} & $I$ & $T^s$ & $I$ & $T^t$
        & R@1 & R@10  & mAP  & R@1 & R@10  & mAP    \\ 
    \hline\hline
    \rowcolor[HTML]{D7F6FF}
    1   &     &     &     & 
        & \textbf{78.82} & 95.47 & 69.68 & \textbf{76.43} & \textbf{94.79} & 67.79 \\
    \hline
    2   & [M] &    &     &                            
        & 78.44 & 95.44 & 69.67 & 76.23 & 94.67 & 67.89  \\
    3   &     &    & [M] & 
        & 78.62 & 95.50 & \textbf{69.87} & 76.40 & 94.64 & \textbf{68.05} \\
    4   & [M] &    & [M] & 
        & 78.23 & \textbf{95.73} & 69.62 & 76.42 & 94.66 & 67.87 \\
    \hline
    5   &     & [M] &     &                            
        & 78.35	& 95.42 & 69.42 & 75.94 & 94.32 & 67.60  \\
    6   &     &     &     & [M]
        & 77.92	& 95.34 & 69.35 & 75.65	& 94.56 & 67.56 \\
    7   &     & [M] &     & [M]
        & 78.64	& 95.48	& 69.53 & 76.01	& 94.35 & 67.71 \\
    \hline\thickhline
    \end{tabular}
    }}
    \label{tab: itc}
\end{table}

\begin{figure*}[tbp]
\includegraphics[width=\textwidth]{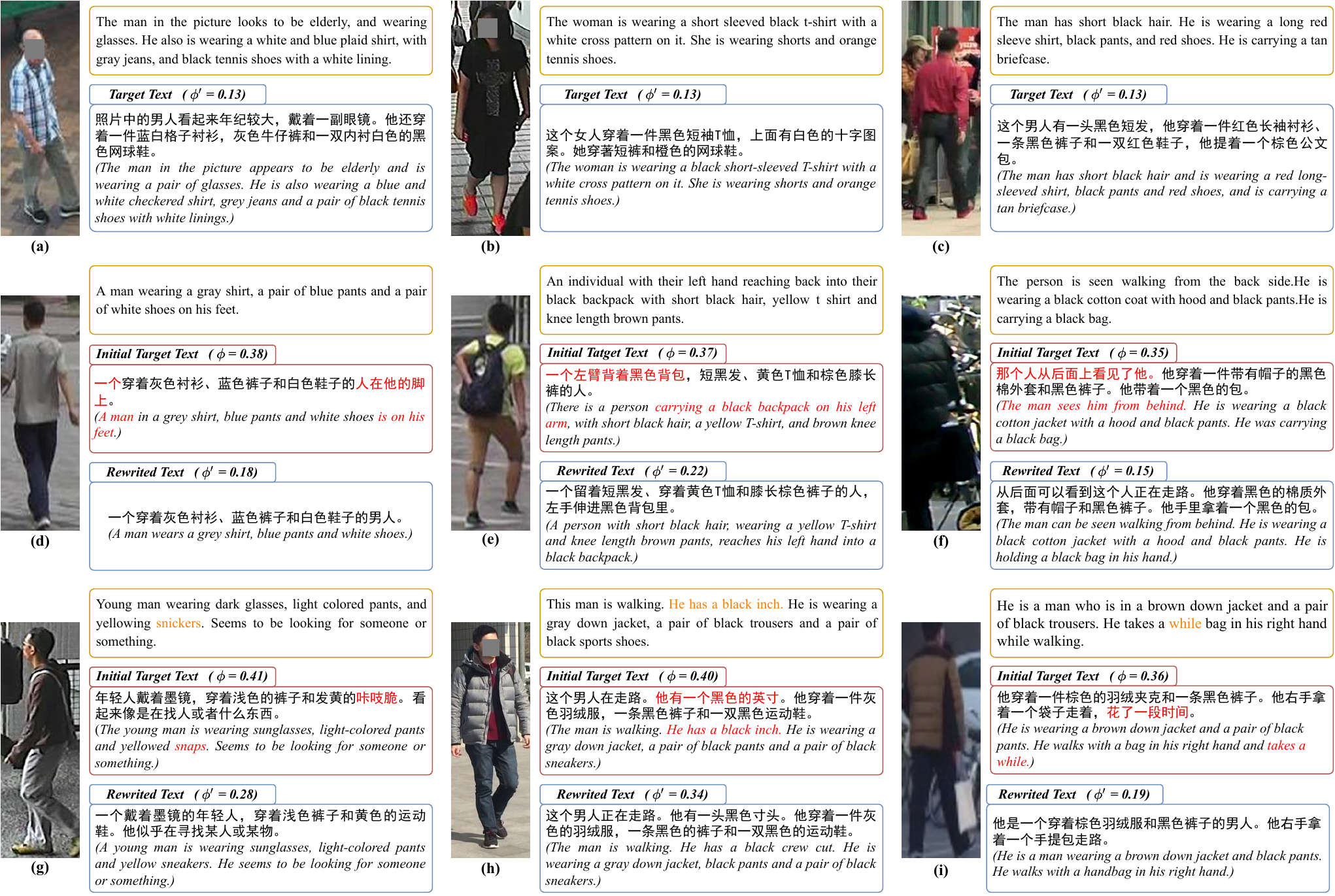}
\caption{Examples of Chinese texts before and after the rewriting phase in LDAT. (a)\(\sim\)(c) display clean initial target texts that do not require rewriting, while (d)\(\sim\)(i) show noisy initial target texts along with their corresponding rewritten versions. Errors in source texts and initial target texts are marked in orange and red, respectively.}
\label{fig: example display}
\end{figure*}

\begin{figure*}[ht]
\includegraphics[width=\textwidth, height=0.27\textwidth]{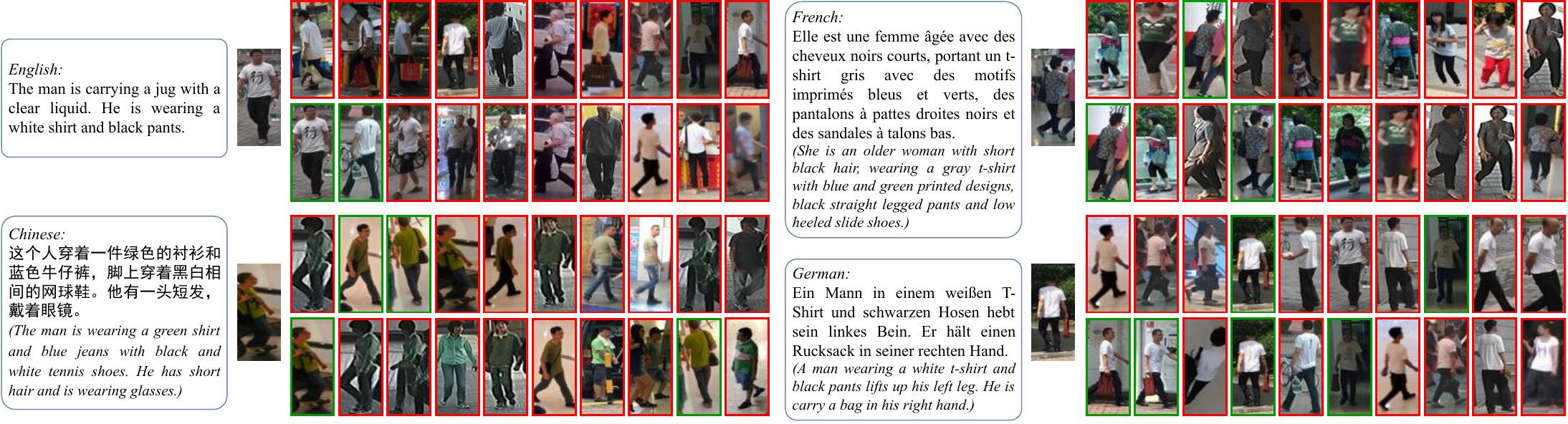}
\caption{Comparison of top-10 retrieved results on CUHK-PEDES(M) between IRRA (the first row) and Bi-IRRA (the second row) for each text query. The matched and mismatched images are marked with green and red rectangles, respectively.}
\label{fig: QR_Bi-IRRA}
\end{figure*}

\subsection{Hyper-parameter analysis}
In this paper, hyper-parameters include the threshold \(\theta\) in Eq.~\eqref{eq: 2} and Eq.~\eqref{eq: 3}, the text mask ratio \(p_{\text{txt}}\), the image mask ratio \(p_{\text{img}}\), and the loss weights \(\lambda_i\) (\(i = 1, 2\)) in Eq.~\eqref{eq: 15}. 

\textbf{The threshold} \(\theta\) determines the division of the clean dataset \(\mathcal{D}^C_{\mathcal{M}}\) and the noisy dataset \(\mathcal{D}^N_{\mathcal{M}}\). 
Taking the initial target texts in Chinese from CUHK-PEDES(M) as the example, we visualize the distribution map of noise levels of all texts in Fig.~\ref{fig: dist_theta} (a). The visualization reveals a denser concentration of texts at low-noise levels compared to high-noise levels, which aligns with expectations: the source texts describing persons typically follow a consistent structure, resulting in relatively accurate translations in the translation phase.

After selecting a specific \(\theta\), these texts are divided into two groups: clean texts (the noise level $\leq$ \(\theta\)) and noisy texts (the noise level $>$ \(\theta\)). Typically, a smaller \(\theta\) results in fewer clean texts, which may hinder the fine-tuning of the MLLM due to insufficient integration of domain knowledge during the rewriting phase. Conversely, a larger \(\theta\) may misclassify more noisy texts as clean, leading to interference during the fine-tuning of the MLLM. These factors collectively influence subsequent retrieval performance. 
In Fig.~\ref{fig: dist_theta} (b), the purple and green bars represent the quantities of clean texts and noisy texts, \(N_{clean}\) and \(N_{noise}\), respectively, while the line chart illustrates the retrieval performance across various \(\theta\) values.
It can be observed that the retrieval performance peaks when \(\theta = 0.15\), indicating a balance between the sizes of clean and noisy data. Consequently, setting \(\theta\) to the rounded mean value of noise levels of all texts proves to be an optimal choice.

\textbf{The mask ratio} \(p_{\text{txt}}\) and \(p_{\text{img}}\) determine the proportion of text and image masked in \emph{bi-lingual MLM} and \emph{cross-lingual D-MIM} tasks, respectively.  
Fig.~\ref{fig: mask ratio} illustrates the impact of different \(p_{\text{txt}}\) and \(p_{\text{img}}\) values on retrieval performance.  
Overall, the performance remains relatively stable across varying values. We empirically set \(p_{\text{txt}} = 0.4\) and \(p_{\text{img}} = 0.5\) for optimal performance.

\textbf{The loss weights} \(\lambda_1\) and \(\lambda_2\) are employed to balance different pretext tasks. We vary their values and report the experimental results in Fig.~\ref{fig: hype}. Overall, adjusting these values shows a relatively stable trend in performance. We set the values as \(\lambda_1 = 4\) and \(\lambda_2 = 4\).

\subsection{Qualitative Results}
\label{sec: qualitative results}

Fig.~\ref{fig: example display} illustrates some examples before and after the rewriting phase in LDAT for generating Chinese texts. 
Fig.~\ref{fig: example display} (a)\(\sim\)(c) showcase clean target texts that do not require rewriting. The original source texts are characterized by the clear and concise descriptions, enabling LLMs to achieve satisfactory translation results. 
Fig.~\ref{fig: example display} (d)\(\sim\)(i) present initial target texts containing noise and their rewritten versions. Translation errors in the initial target texts are highlighted in red. 
Notably, in (h)\(\sim\)(i), noise present in the source texts, marked in orange, significantly misleads LLMs for translation. For instance, the misspelling of \textit{``sneakers''} as \textit{"snickers"} in the source text results in an inaccurate translation in (h). 
These initial target texts are effectively revised during the subsequent rewriting phase. When incorporating the domain knowledge, LDAT demonstrates strong performance in translation, consistently producing accurate target texts. 

Fig.~\ref{fig: QR_Bi-IRRA} presents a comparison of the top-\(10\) retrieval results obtained from IRRA and our proposed Bi-IRRA, utilizing text queries in various languages. As illustrated in the figure, Bi-IRRA demonstrates superior performance in capturing fine-grained information such as \textit{``carrying a jug
with a clear liquid''}, \textit{``white shirt''}, and \textit{``black pants''}, leading to significantly more accurate retrieval results with queries in multiple languages. 

\section{Conclusion}
This paper pioneers a multilingual TIPR task, exploring TIPR in multilingual scenarios. 
First, we propose the LDAT pipeline to construct a multilingual TIPR benchmark automatically. LDAT alleviates noise issues in large model translations by effectively acquiring and leveraging domain-specific knowledge, enabling the efficient creation of high-quality multilingual TIPR datasets.  
In addition, we introduce Bi-IRRA: a cross-modal Bidirectional Implicit Relation Reasoning and Aligning framework to achieve comprehensive alignment across different languages and modalities.
Extensive experiments demonstrate that the proposed framework consistently achieves superior retrieval performance across various languages. We believe that research on the multilingual TIPR task can further drive the practical application of this field.


\bibliographystyle{IEEEtran}
\bibliography{main}

\newpage

\begin{IEEEbiography}
[{\includegraphics[width=1in, height=1.25in, clip, keepaspectratio]{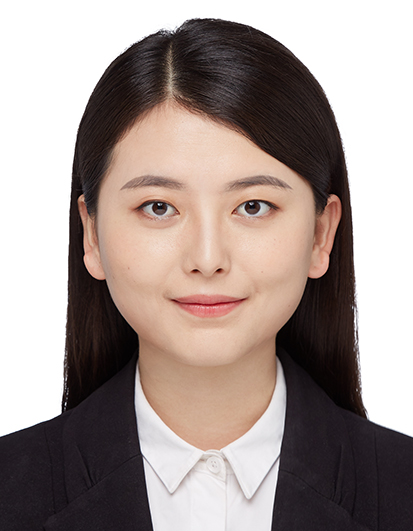}}]
{Min Cao} received her Ph.D. degree in pattern recognition and intelligent systems from the Institute of Automation, Chinese Academy of Sciences, Beijing, China, in 2020. In March 2020, she became a member of the computer science and technology school at Soochow University, where she is currently an Associate Professor. She was a visiting scholar in computer graphics research, Fraunhofer-Gesellschaft, Darmstadt, German, in 2018. Her research interests include cross-modal vision-language learning and person re-identification.
\end{IEEEbiography}

\vspace{-1cm}
\begin{IEEEbiography}
[{\includegraphics[width=1in, height=1.25in, clip, keepaspectratio]{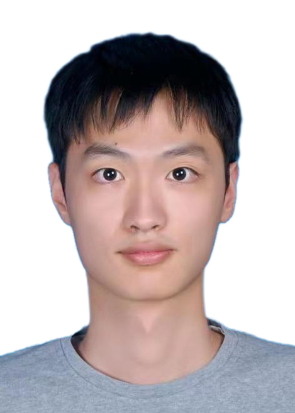}}]
{Xinyu Zhou} received the BE degree from the School of Computer Science and Technology, Soochow University, Suzhou, China. He is currently work toward the ME degree with the School of Computer Science and Technology, Soochow University. His research interests include cross-modal retrieval and person re-identification.
\end{IEEEbiography}

\vspace{-1cm}
\begin{IEEEbiography}
[{\includegraphics[width=1in, height=1.25in, clip, keepaspectratio]{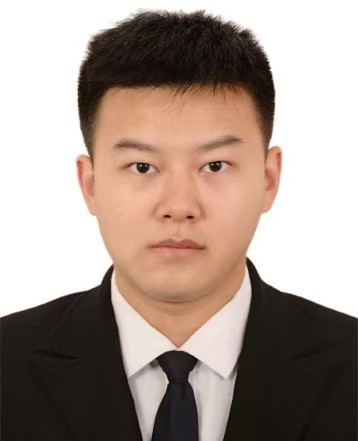}}]
{Ding Jiang} received the Master's degree from the School of Computer Science, Wuhan University, Wuhan, China. His research interests include cross-modal retrieval, person re-identification and 3D human pose estimation.
\end{IEEEbiography}

\vspace{-1cm}
\begin{IEEEbiography}
[{\includegraphics[width=1in, height=1.25in, clip, keepaspectratio]{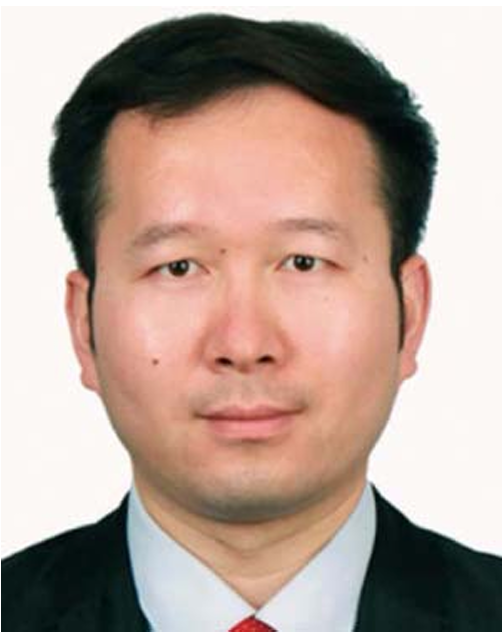}}]
{Bo Du} (Senior Member, IEEE) received the PhD degree in photogrammetry and remote sensing from the State Key Laboratory of Information Engineering in Surveying, Mapping and Remote Sensing, Wuhan University, Wuhan, China, in 2010. He is a professor with the School of Computer Science, Wuhan University. He has more than 60 research articles published in the IEEE TGRS, TIP, JSTARS, and GRSL. Thirteen of them are ESI hot articles or highly cited articles. His major research interests include pattern recognition, hyperspectral image processing, machine learning, and signal processing. He was a recipient of the Distinguished Paper Award from IJCAI 2018, the Best Paper Award of the IEEE Whispers 2018, the Champion Award of the IEEE Data Fusion Contest 2018, the Best Reviewer Award from the IEEE GRSS for his service to the IEEE Journal of Selected Topics in Earth Observations and Applied Remote Sensing, in 2011, and the ACM rising star awards for his academic progress, in 2015. He was the session chair of the IGARSS 2018/2016 and the 4th IEEE GRSS Workshop on Hyperspectral Image and Signal Processing: Evolution in Remote Sensing.
\end{IEEEbiography}

\vspace{-1cm}
\begin{IEEEbiography}
[{\includegraphics[width=1in, height=1.25in, clip, keepaspectratio]{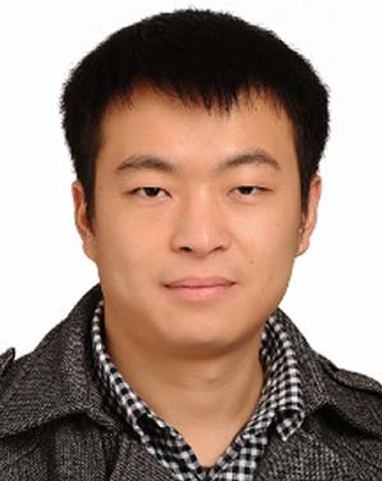}}]
{Mang Ye} (Senior Member, IEEE) received the PhD degree in computer science from Hong Kong Baptist University, in 2019. He is currently a full professor with the School of Computer Science, Wuhan University, Wuhan, China. He has published more than 100 articles in top-tier venues. He serves as the associate editor for IEEE Transactions on Image Processing, IEEE Transactions on Information Forensics and Security, the Journal of Electronic Imaging, and CAAI Transactions on Intelligence Technology. His research interests focus on computer vision, pattern recognition, and federated learning.
\end{IEEEbiography}

\begin{IEEEbiography}
[{\includegraphics[width=1in, height=1.25in, clip, keepaspectratio]{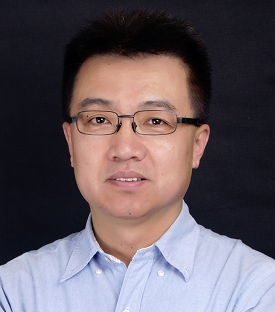}}]
{Min Zhang} (Fellow, ACL) received the bachelor’s and PhD degrees from the Harbin Institute of Technology, in 1991 and 1997, respectively. He is a distinguished professor with Soochow University (China). His current research interests include machine translation, natural language processing, large model and AI.
\end{IEEEbiography}

\end{document}